%% file: main.tex
\documentclass{article}

\title{\huge {HyenaDNA: Long-Range Genomic Sequence} \\ Modeling at  Single Nucleotide Resolution}

\author{Eric Nguyen\footnote{Equal contribution. $\dagger$ Equal senior authorship. $^1$Stanford University. $^2$Harvard University. $^3$SynTensor. $^4$Mila and Universit\'e de Montr\'eal.}~$^{,1}$, Michael Poli$^{*,1}$, Marjan Faizi$^{2,*}$, \\ Armin W. Thomas$^1$, Callum Birch Sykes$^3$, Michael Wornow$^1$, Aman Patel$^1$, \\ 
Clayton Rabideau$^3$, Stefano Massaroli$^4$, Yoshua Bengio$^4$, Stefano Ermon$^1$, \\
Stephen A. Baccus$^{1,\dagger}$, Christopher R\'e$^{1,\dagger}$
}

\input{_template/style}
\input{_template/math}

\numberwithin{equation}{section}

\begin{document}
\maketitle

\input{hyenadna/0_abstract}

\doparttoc

%
\input{hyenadna/1_intro}

\input{hyenadna/2_background}

\input{hyenadna/3_hyena.tex}

\input{hyenadna/4_experiments}
\input{hyenadna/5_conclusion}

\input{hyenadna/6_ack}

\bibliographystyle{abbrvnat}
\bibliography{_bibliography/main.bib}
\newpage
\appendix
\rule[0pt]{\columnwidth}{1pt}
\begin{center}
    \huge{HyenaDNA} \\
    \vspace{0.3cm}
    \emph{Supplementary Material}
\end{center}
\rule[0pt]{\columnwidth}{1.5pt}
\doparttoc
\tableofcontents
\newpage

\input{hyenadna/appendix/experiments}

\end{document}

%% file: _template/style.tex
\usepackage{lipsum}
\usepackage[utf8]{inputenc}
\usepackage[T1]{fontenc}    
\usepackage[sf]{noto}
\usepackage[bookmarks=true, colorlinks=true,linkcolor=blue!70,citecolor=green!80!black]{hyperref}
\usepackage[round]{natbib}
\usepackage{wrapfig}
\usepackage{booktabs}  
\usepackage{longtable}
\usepackage{multirow}
\usepackage{ducksay}
\usepackage{etaremune}
\usepackage{afterpage}
\usepackage{capt-of}
\usepackage[table, x11names]{xcolor}
\usepackage{decorule}
\usepackage{scalerel,xparse}
\usepackage{enumitem} 
\usepackage[top=25mm, bottom=25mm, left=25mm, right=25mm,
            foot=5mm, marginparsep=0mm]{geometry}
\usepackage{svg}

\usepackage{chngcntr}
\counterwithin{figure}{section} 
\counterwithin{table}{section}

\usepackage{graphicx}
\usepackage{tikz}
\usepackage{tikz-cd}
\usepackage{hf-tikz}
\usepackage{pgfplots} 
\pgfplotsset{compat=1.17} 
\pgfplotsset{
        table/search path={figures/drawings},
    }
\usetikzlibrary{fadings}
\usetikzlibrary{shapes, arrows, fit, backgrounds, arrows.meta}

\usetikzlibrary{matrix}
\usetikzlibrary{shadows.blur}
\usetikzlibrary{patterns, tikzmark}
\usetikzlibrary{decorations.pathreplacing, calc, decorations.markings,}
\usetikzlibrary{positioning}

\usepgfplotslibrary{groupplots}
\usepgfplotslibrary{patchplots}

\definecolor{bg}{gray}{0.97}
\definecolor{olive}{rgb}{0.6, 0.6, 0.2}
\definecolor{sand}{rgb}{0.8666666666666667, 0.8, 0.4666666666666667}
\definecolor{wine}{rgb}{0.5333333333333333, 0.13333333333333333, 0.3333333333333333}
\definecolor{deblue}{RGB}{11,132,147}
\definecolor{ocra}{RGB}{204, 119, 34}
\definecolor{emph_color}{RGB}{12, 100, 210}

\definecolor{darkersteelblue}{RGB}{25, 89, 140}

\usepackage{lettrine}
\usepackage{Typocaps}

\LettrineTextFont{\itshape}
\usepackage{amsthm}

\newtheorem{proposition}{Proposition}[section]

\usepackage{minitoc}

\newcommand{\chapref}[1]{\hyperref[#1]{Chapter \ref{#1}}}
\newcommand{\secref}[1]{\hyperref[#1]{Section \ref{#1}}}
\usepackage{marginnote}

\usepackage[many]{tcolorbox}
\tcbuselibrary{breakable,xparse,skins}

\usepackage{newunicodechar}
\newunicodechar{λ}{{$\mathtt\lambda$}}
\newunicodechar{μ}{{$\mathtt\mu$}}

\usepackage{xparse}
\DeclareTColorBox{emphbox}{O{black}O{0cm}}{
    enhanced jigsaw,
    breakable,
    outer arc=0pt,
    arc=0pt,
    colback=white,
    rightrule=0pt,
    toprule=0pt,
    top=0pt,
    right=0pt,
    bottom=0pt,
    bottomrule=0pt,
    colframe=#1,
    enlarge left by=#2,
    width=\linewidth-#2,
}

\DeclareTColorBox{emphbox}{O{black}O{0cm}}{
    empty,
    breakable=true,
    outer arc=0pt,
    arc=0pt,
    rightrule=0pt,
    leftrule=2pt,
    borderline west={2pt}{0pt}{#1},
    toprule=0pt,
    top=0pt,
    right=-3pt,
    bottom=0pt,
    bottomrule=0pt,
    colframe=#1,
    enlarge left by=#2,
    width=\linewidth-#2,
}
\usepackage{listings}
\makeatletter
\newcommand\BeraMonottfamily{%
  \def\fvm@Scale{0.85}
  \fontfamily{fvm}\selectfont
}
\makeatother

\usepackage{pifont}
\newcommand{\xmark}{\ding{55}}%


%% file: _template/math.tex
\usepackage{amsmath, amssymb, amsfonts, mathtools}
\usepackage{physics}
\usepackage{cancel}
\usepackage{thmtools, thm-restate}
\usepackage{accents}
\usepackage{bm}

\makeatletter
\newcommand{\ostar}{\mathbin{\mathpalette\make@circled *}}
\newcommand{\make@circled}[2]{%
  \ooalign{$\m@th#1\smallbigcirc{#1}$\cr\hidewidth$\m@th#1#2$\hidewidth\cr}%
}
\newcommand{\smallbigcirc}[1]{%
  \vcenter{\hbox{\scalebox{0.77778}{$\m@th#1\bigcirc$}}}%
}
\makeatother

\newcommand{\x}{\times}

\usepackage{pict2e, picture}
\makeatletter
\DeclareRobustCommand{\Arrow}[1][]{%
\check@mathfonts
\if\relax\detokenize{#1}\relax
\settowidth{\dimen@}{$\m@th\rightarrow$}%
\else
\setlength{\dimen@}{#1}%
\fi
\sbox\z@{\usefont{U}{lasy}{m}{n}\symbol{41}}%
\begin{picture}(\dimen@,\ht\z@)
\roundcap
\put(\dimexpr\dimen@-.7\wd\z@,0){\usebox\z@}
\put(0,\fontdimen22\textfont2){\line(1,0){\dimen@}}
\end{picture}%
}
\makeatother

\usepackage{scalerel}
\newcommand{\cO}{\mathcal{O}}

\newcommand{\sA}{\mathsf{A}}

\newcommand{\sD}{\mathsf{D}}

\newcommand{\sH}{\mathsf{H}}

\newcommand{\sT}{\mathsf{T}}

\newcommand{\sW}{\mathsf{W}}

\newcommand{\R}{\mathbb{R}}

\DeclareMathAlphabet{\nummathbb}{U}{BOONDOX-ds}{m}{n}

\makeatletter
\DeclareRobustCommand\widecheck[1]{{\mathpalette\@widecheck{#1}}}
\def\@widecheck#1#2{%
    \setbox\z@\hbox{\m@th$#1#2$}%
    \setbox\tw@\hbox{\m@th$#1%
       \widehat{%
          \vrule\@width\z@\@height\ht\z@
          \vrule\@height\z@\@width\wd\z@}$}%
    \dp\tw@-\ht\z@
    \@tempdima\ht\z@ \advance\@tempdima2\ht\tw@ \divide\@tempdima\thr@@
    \setbox\tw@\hbox{%
       \raise\@tempdima\hbox{\scalebox{1}[-1]{\lower\@tempdima\box
\tw@}}}%
    {\ooalign{\box\tw@ \cr \box\z@}}}
\makeatother

%% file: hyenadna/0_abstract.tex
\begin{abstract}

Genomic (DNA) sequences encode an enormous amount of information for gene regulation, protein synthesis, and numerous other cellular properties.
Similar to natural language models, researchers have proposed foundation models in genomics to learn generalizable features from unlabeled genome data that can then be fine-tuned for downstream tasks such as identifying regulatory elements.
Due to the quadratic scaling of attention, previous Transformer-based genomic models have used 512 to 4k tokens as context (<0.001\% of the human genome), significantly limiting the modeling of long-range interactions in DNA. 
In addition, these methods rely on tokenizers or fixed k-mers to aggregate meaningful DNA units, losing single nucleotide resolution (i.e. DNA "characters") where subtle genetic variations can completely alter protein function via single nucleotide polymorphisms (SNPs).
Recently, {$\sf Hyena$}, a large language model based on implicit convolutions was shown to match attention in quality while allowing longer context lengths and lower time complexity.
Leveraging {$\sf Hyena$}’s new long-range capabilities, we present {$\sf HyenaDNA$}, \textbf{a genomic foundation model} pretrained on the human reference genome with \textbf{context lengths of up to 1 million tokens at the single nucleotide-level} – an \textbf{up to 500x increase} over previous dense attention-based models.
{$\sf HyenaDNA$} scales sub-quadratically in sequence length (training up to 160x faster than Transformer), \textbf{uses single nucleotide tokens}, and has \textbf{full global context at each layer}.
We explore what longer context enables - including the first use of in-context learning in genomics for simple adaptation to novel tasks without updating pretrained model weights.
On a long-range species classification task, {$\sf HyenaDNA$} is able to effectively solve the challenge by increasing the context length to 1M without downsampling.
On fine-tuned benchmarks from the Nucleotide Transformer, {$\sf HyenaDNA$} reaches state-of-the-art (SotA) on 12 of 18 datasets using a model with orders of magnitude less parameters and pretraining data.\footnote{On benchmarks from Nucleotide Transformer, {$\sf HyenaDNA$} uses a model with 1500x fewer parameters (2.5B vs 1.6M) and 3200x less pretraining data (3202 vs 1 human reference genome).}
On the GenomicBenchmarks, {$\sf HyenaDNA$} surpasses SotA on 7 of 8 datasets on average by +10 accuracy points, and by as much as +20 accuracy points on enhancer identification.
Code available at \href{URL}{https://github.com/HazyResearch/hyena-dna}.
\end{abstract}

%% file: hyenadna/1_intro.tex
\section{Introduction}

Understanding and learning from DNA sequences has long been a goal of biologists and deep learning researchers, as its “language” encodes instructions essential for all living things \citep{encode2020}. The mapping from DNA instructions, genotypes, to observable function and traits, phenotypes, remains on-going research effort. Towards this goal, researchers have proposed using foundation models (FMs) in genomics to learn generalizable features from unstructured whole genome data that can then be fine-tuned for a number of tasks including predicting the location and function of genes, identifying regulatory elements, and analyzing the evolution of species \citep{ji2021dnabert, dallatorre2023nucleotide, gankin2023species, benegas2022dna, yang2022scbert, zvyagin2022genslms}. In contrast to protein sequences, which have had successes in protein language models \citep{lin2022language, madani2023large, meier2021language, ferruz2022protgpt2, brandes2022proteinbert,rao2020transformer,elnaggar2021prottrans}, DNA sequences are orders of magnitudes longer (e.g. the human genome is 3.2B nucleotides) with long-range dependencies and interactions that span over 100k+ nucleotides in length \citep{avsec2021effective}. Overcoming the long-range limitations of current generation models could help drive the next wave of innovations in AI-powered drug discovery and therapeutics, and enable genomic FMs to understand and learn in-context whole patient genomes in a personalized way.

\begin{figure}[t]
    \centering
    \includegraphics[width=\linewidth]{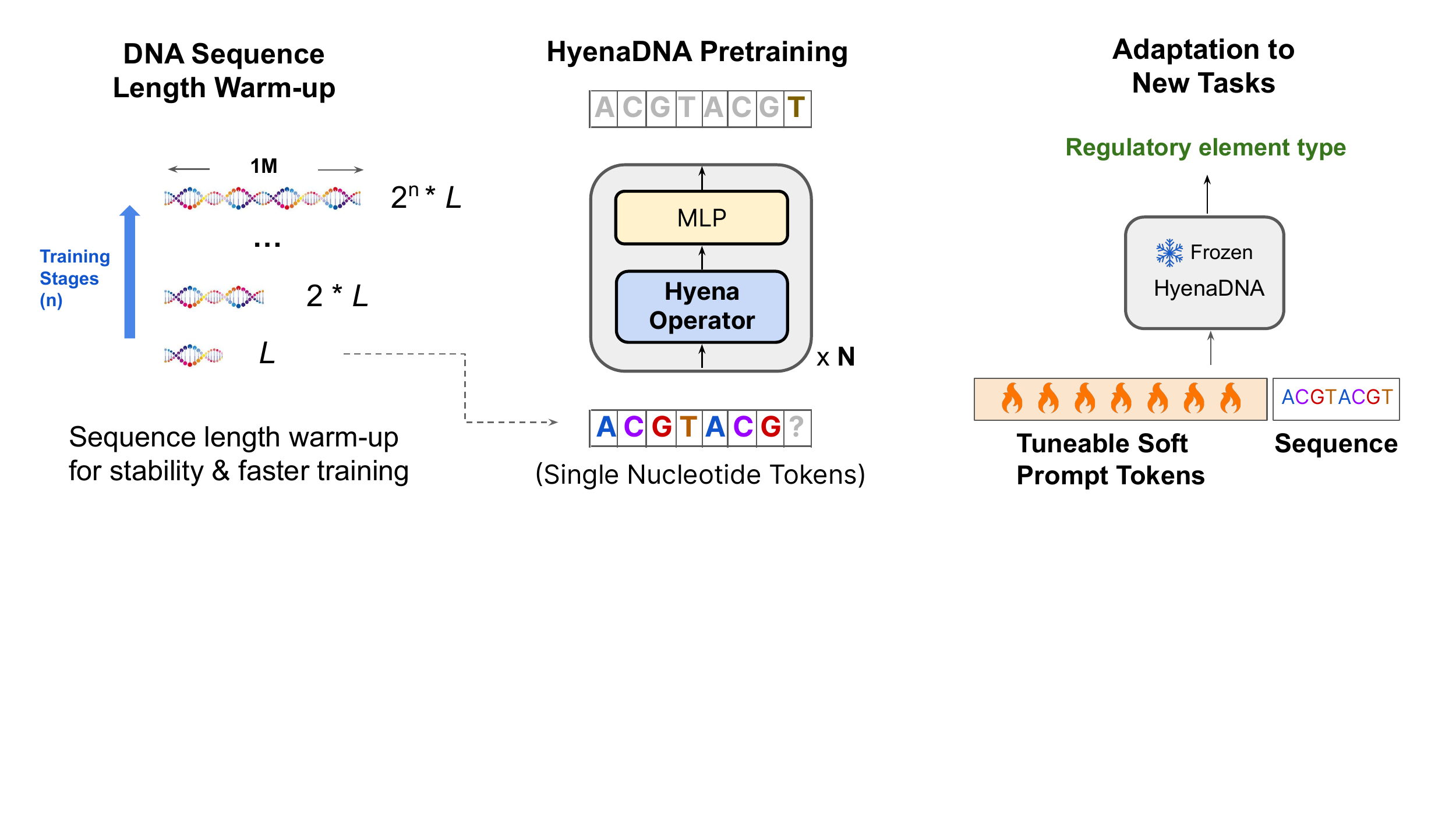}
    \vspace{-37mm}
    \caption{{$\sf HyenaDNA$} recipe for long-range foundation models in genomics. The {$\sf HyenaDNA$} architecture is a simple stack of {$\sf Hyena$} operators \citep{poli2023hyena} trained using next token prediction. (See Fig. \ref{fig:hyena-arch} for block diagram of architecture). We introduce a new sequence length scheduling technique to stabilize training, and provide a method to leverage the longer context length to adapt to novel tasks without standard fine-tuning by filling the context window with learnable soft prompt tokens.}
    \label{fig:fullstack}
\end{figure}

\begin{wrapfigure}[18]{r}{0.5\linewidth}
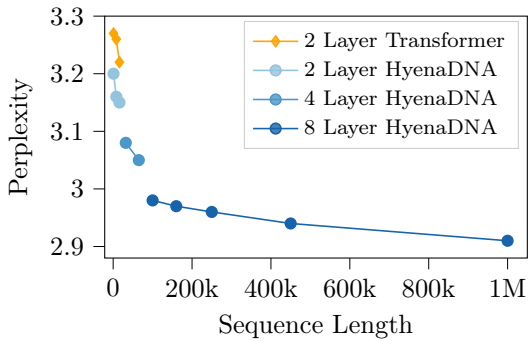

    \vspace{-5mm}
    \centering
    \include{figs/source/ppl}
    \vspace{-5mm}
    \caption{Pretraining on the human reference genome using longer sequences leads to better perplexity (improved prediction of next token).}
    \label{fig:ppl_seqlen}
\end{wrapfigure}

\paragraph{Limitations of current models}
Previous genomic FM approaches have relied on attention-based Transformers \citep{ji2021dnabert, dallatorre2023nucleotide, yang2022scbert, zvyagin2022genslms}, but face a number of challenges unique to DNA sequences. The attention mechanism scales quadratically in sequence length, with current genomic FMs pretraining on only 512 to 4,096 tokens as context \citep{ji2021dnabert, zvyagin2022genslms, dallatorre2023nucleotide, zaheer2020big}, <0.001\% of the human genome.
Also prevalent is the reliance on fixed k-mers, akin to DNA “words”, and tokenizers to aggregate meaningful DNA units. However, single nucleotide alterations represent physical analogs where, for example, single nucleotide polymorphisms (SNPs) and mutations can have a profound impact on biological properties including regulatory activity \citep{nasser2021}. In contrast, natural language semantics can often be conserved when single character or word changes occur over very long contexts. Therefore, having both \textbf{long-range context} and \textbf{single nucleotide resolution} simultaneously is critical, and remains a particular challenge in genomics.

\begin{figure}[h]
    \centering
    \includegraphics[width=0.9\linewidth]{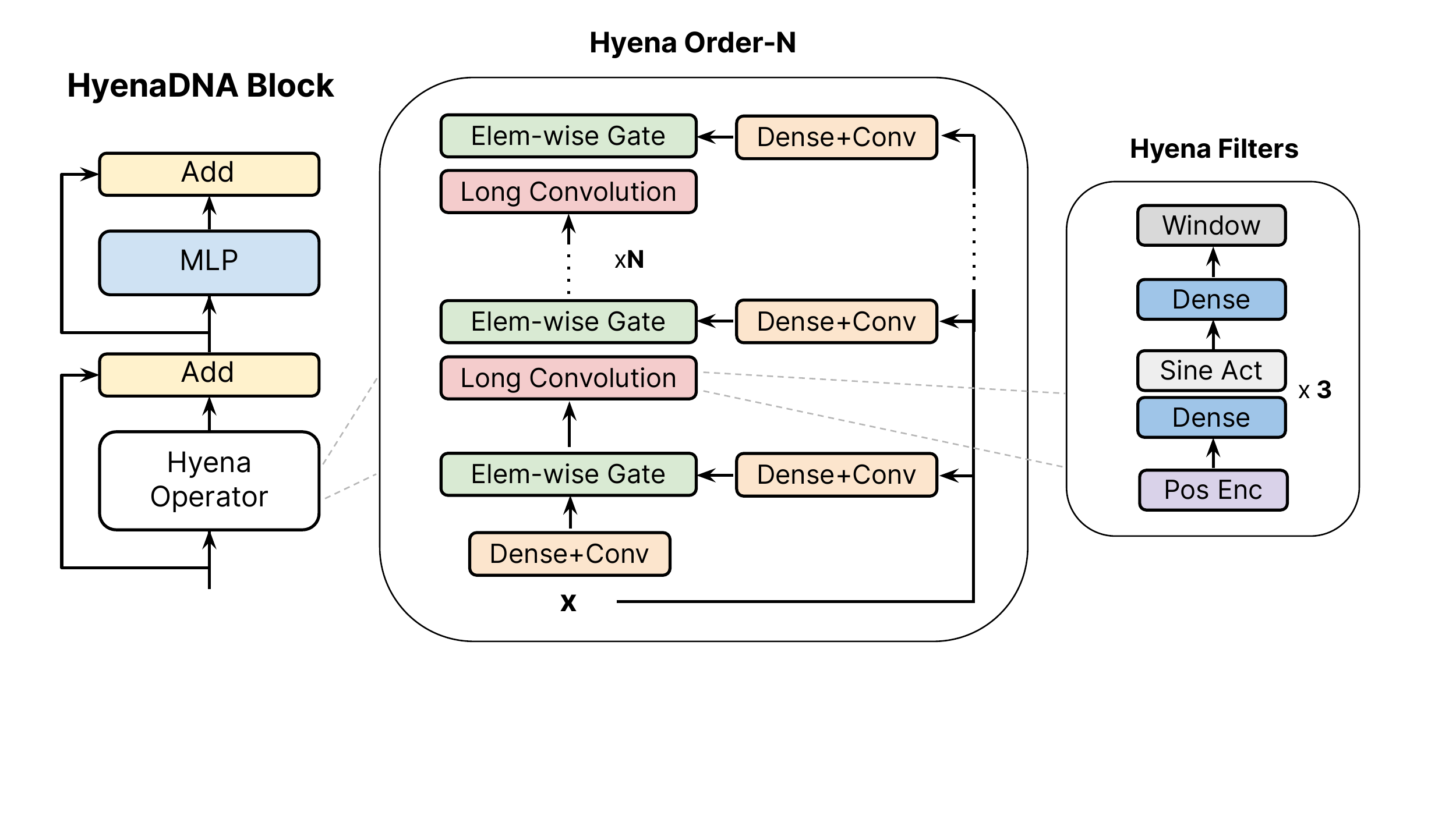}
    \vspace{-14mm}
    \caption{{$\sf HyenaDNA$} block architecture. A Hyena operator is composed of long convolutions and element-wise gate layers. The gates are fed projections of the input using dense layers and short convolutions. The long convolutions are parameterized \textit{implicitly} via an MLP that produces the convolutional filters. The convolution itself is evaluated using a Fast Fourier Transform convolution with time complexity $\cO(L \log_2 L)$.}
\label{fig:hyena-arch}
\end{figure}

\paragraph{Toward longer context models}
Recently, {$\sf Hyena$} \citep{poli2023hyena}, a large language model based on implicit convolutions, was shown to match attention in quality while reducing computational time complexity, thereby allowing a longer context to be processed.
Hyena uses a parameter-efficient \textbf{global convolutional filter} along with a \textbf{data-controlled gating} mechanism, which enables a context-specific operation over every token. 
Indeed, {$\sf Hyena$} showed that for simple associative recall tasks using \textit{synthetic} data, a shallow 2 layer model could effectively process context lengths at 131k tokens. We hypothesize that ${\sf Hyena}$’s core operations can unlock the potential to capture both the long-range and single nucleotide resolution of \textit{real} genomic sequences over attention-based approaches. To test this, we explore two questions: \textbf{(i.) Can a convolutional long-context model be used effectively at single nucleotide resolution? (ii.) What new capabilities could long-context genomic foundations models enable?}

\paragraph{{$\sf HyenaDNA$}}
The result of our investigation is {$\sf HyenaDNA$}, a genomic FM pretrained on the human reference genome at \textbf{context lengths up to 1 million tokens at single nucleotide resolution} - an up to \textbf{500x increase} over existing genomic FMs using dense-attention. {$\sf HyenaDNA$} scales sub-quadratically in sequence length (training up to 160x faster than attention at sequence length 1M), uses single nucleotide tokens, and has a global receptive field at each layer. Our contributions include a "full-stack" recipe for building genomic FMs, including architecture design, a warm-up schedule to speed up training on ultralong sequences, and an efficient downstream adaptation procedure based on soft prompting and in-context learning. 

\paragraph{Full-stack genomics modeling} 

We start with a decoder-only Hyena architecture pretrained using next nucleotide (token) prediction. We forego standard aggregating tokenizers, using a single-character tokenizer and a minimal DNA vocabulary of 4 nucleotides (plus special tokens). Training stability becomes an issue at ultralong sequences (200k+). To overcome this issue, we introduce a sequence length warm-up scheduler that gradually increases sequence length in stages. At sequence length 450k, training time is reduced by 40\%, while boosting accuracy by 7.5 accuracy points on a species classification task. Furthermore, we design downstream adaptation procedures to leverage longer context windows, as simpler and more flexible alternatives to standard fine-tuning in genomics. This includes a novel soft prompt technique where learnable tokens (up to 32k) are injected directly into the input sequence itself, enabling competitive downstream results without the need to update a pretrained model. 

\paragraph{Genomic downstream tasks}
We apply our pretrained {$\sf HyenaDNA$} models to 29 diverse downstream genomic tasks to showcase its long-range ability as well as fine-grain resolution. 
On fine-tuned benchmarks from the Nucleotide Transformer \citep{dallatorre2023nucleotide}, {$\sf HyenaDNA$} achieves state-of-the-art (SotA) on 12 of 18 datasets while using a model with orders of magnitude less parameters and pretraining data (see Tab. \ref{tab:nuctran}).
On the GenomicBenchmarks \citep{gresova2022genomic}, {$\sf HyenaDNA$} surpasses SotA on 7 of 8 datasets on average by +10 accuracy points, and by as much as +20 accuracy points on enhancer function identification.
On a novel species classification task, {$\sf HyenaDNA$} effectively solves the challenge by increasing the context length to 1 million tokens.
In a challenging chromatin profile experiment, a 919-way multi-task, {$\sf HyenaDNA$} performs competitively against a larger SotA sparse-attention BigBird Transformer \citep{zaheer2020big}.
Finally, we analyze the learned embeddings of a pretrained {$\sf HyenaDNA$} model by clustering sequences by biotype (gene or transcription type) and compare the results with existing genomic FMs, showing that {$\sf HyenaDNA$} can serve as an effective universal featurizer in genomics.

%% file: figs/source/ppl.tex
\begin{minipage}{\textwidth}
\begin{tikzpicture}

\definecolor{darkcyan23100171}{RGB}{23,100,171}
\definecolor{darkgray176}{RGB}{176,176,176}
\definecolor{lightgray204}{RGB}{204,204,204}
\definecolor{orange}{RGB}{255,165,0}
\definecolor{skyblue147196222}{RGB}{147,196,222}
\definecolor{steelblue74151201}{RGB}{74,151,201}

\begin{axis}[
height=4.8cm,
width=7.2cm,
legend cell align={left},
legend style={fill opacity=0.8, draw opacity=1, text opacity=1, draw=lightgray204, font=\footnotesize},
tick align=outside,
tick pos=left,
title={\textbf{PPL vs Context on the Human Genome}},
x grid style={darkgray176},
title style={yshift=2mm},
xlabel={Sequence Length},
xticklabels={0, 0,200k,400k,600k,800k,1M},
xmin=-21.45, xmax=1050,
xtick style={color=black},
y grid style={darkgray176},
ylabel={Perplexity},
ymin=2.88, ymax=3.30,
ytick style={color=black}
]
\addplot [semithick, orange, mark=diamond*, mark size=2, mark options={solid}]
table {%
1 3.27
8 3.26
16 3.22
};
\addlegendentry{2 Layer Transformer}
\addplot [semithick, skyblue147196222, mark=*, mark size=2, mark options={solid}]
table {%
1 3.2
8 3.16
16 3.15
};
\addlegendentry{2 Layer HyenaDNA}
\addplot [semithick, steelblue74151201, mark=*, mark size=2, mark options={solid}]
table {%
32 3.08
65 3.05
};
\addlegendentry{4 Layer HyenaDNA}
\addplot [semithick, darkcyan23100171, mark=*, mark size=2, mark options={solid}]
table {%
100 2.98
160 2.97
250 2.96
450 2.94
1000 2.91
};
\addlegendentry{8 Layer HyenaDNA}
\end{axis}

\end{tikzpicture}

\end{minipage}

%% file: hyenadna/2_background.tex
\section{Preliminaries and Related Work}

\subsection{Transformers and Attention}
Powering many recent \textit{foundation models} is the \textit{attention} mechanism. Given a length-$L$ sequence $x\in\R^{L\times D}$, a (single-headed) layer of \textit{scaled self-attention} \citep{bahdanau2014neural, vaswani2017attention} is a map from $\R^{L\times D}$ to $\R^{L\times D}$ which performs the following operations:
\begin{equation}\label{eq:att}
    \begin{aligned}
    \sA(x) = \sigma( x \sW_q \sW_k^\top x^\top), \quad y = \sA(x) x \sW_v\\
    \end{aligned}
\end{equation}
where $D$ is the embedding dimension, $\sW_q, \sW_k, \sW_v\in\R^{D\x D}$ are learnable linear maps and $\sigma$ indicated row-wise softmax (and optional scaling). Attention computes all pair-wise comparison for every token, and scales as $\mathcal{O}(L^2)$ in sequence length. This allows a global context at high resolution, but limits the size of the context on current hardware.

Previous methods to reduce the quadratic cost of attention have used specialized methods to approximate full dense attention \citep{fournier2021practical}. In sparse attention, elements attend only to a subset of all other positions. Alternatively, linear attention methods construct approximations to $\sA(u)$ that can be evaluated in subquadratic time. Both of these classes of methods, however, trade lower time complexity (allowing longer sequences) for loss in expressivity.

\subsection{Long Context Strategies in Genomics}

To achieve longer context, genomic models have relied on two strategies: i. tokenization and ii. dilation and downsampling. Tokenization is a necessary step in masked language modeling (MLM) with bidirectional Transformer architectures (BERT) \citep{devlin2018bert}, a common model in genomics. These tokenizers use fixed k-mers (short overlapping sequences of length k) or frequency-based byte pair encoding (BPE), that attempt to aggregate DNA into meaningful units \citep{ji2021dnabert, zaheer2020big}. Consequently, these aggregation techniques create large new vocabularies (compared to the natural vocabulary of 4 nucleotides) that are less generalizable \citep{tay2021charformer}. The second strategy uses dilated convolutions and downsampling, both of which essentially average or skip elements between weights \citep{fournier2021practical}. A canonical example is the Enformer, which uses dilation and downsampling to reach context lengths of 100k nucleotides to predict gene expression tracks \citep{avsec2021effective}. Common across tokenization, dilation, and downsampling is the sacrifice of single nucleotide resolution to reach longer context.

\subsection{Large Convolutional Models}

A discrete convolution between an input $x$ of length $L$ and a (learnable) filter $h$ is given by: 
\begin{equation}\label{eq:cnn}
    \begin{aligned} 
        y_t = (h * x)_t = \sum_{t'=0}^{L-1} h_{t - t'} x_{t'} \quad \text{or equivalently} \quad y = \sT x. 
    \end{aligned}
\end{equation}
where $\sT\in\R^{L\times L}$ is the Toeplitz matrix corresponding to the convolution. Historically, convolutions have played an important role in deep learning and more broadly signal processing. More recently, it has been shown that by stacking $k$ long convolution layers, where $k$ is parametrized through a function $\gamma_\theta$ i.e. $k := \gamma_\theta(L)$, one can achieve state-of-the-art performance on a variety of benchmarks involving long sequences, for example the Long Range Arena (LRA) \citep{tay2020long,gu2021efficiently,smith2022simplified,fu2023simple}. Different $\gamma_\theta$ have been proposed in the literature: state-space models \citep{gu2021efficiently,fu2023simple}, and implicit parametrizations via neural fields \citep{romero2021ckconv,romero2021flexconv,poli2023hyena}. On language, the {\sf H}-family of implicit convolution language models, {$\sf H3$} and {$\sf Hyena$}, \citep{dao2022hungry,poli2023hyena} used 
long convolutions and gating to match Transformer performance in $\mathcal{O}(L \log_2 L)$ time, notably lower than the $\mathcal{O}(L^2)$ of attention-based models. 

{$\sf HyenaDNA$} takes inspiration from these approaches, showing that attention-free, long-context causal models can achieve high performance on downstream genomic tasks. These extended long-range capabilities enable us to explore new paradigms in genomics, such as in-context learning to easily adapt to new tasks without updating pretrained models.

%% file: hyenadna/3_hyena.tex
\section{\texorpdfstring{${\sf HyenaDNA}$}: Long-Range Genomic Foundation Models}

In this section, we introduce the {\sf HyenaDNA} approach to long-range genomic sequence modeling. We start with a description of the model architecture, then discuss sequence length warm-up and soft prompting techniques for downstream adaptation.

\subsection{The {\sf HyenaDNA} Model}

The ${\sf HyenaDNA}$ model is a decoder-only, sequence-to-sequence architecture defined by a stack of blocks consisting of a ${\sf Hyena}$ operator \citep{poli2023hyena}, followed by a feed-forward neural network (see Fig. \ref{fig:hyena-arch}).

\begin{wrapfigure}[11]{r}{0.45\textwidth}
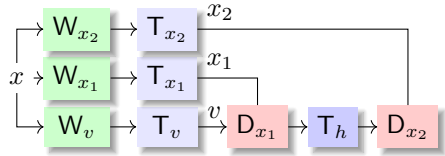
  
    \vspace{-4mm}
    \centering
    \include{figs/source/hyena_operator}
    \vspace{-2mm}
    \caption{The ${\sf Hyena}$ operator is a combination of long convolutions $\sT$ and data-controlled gating $\sD$, and can be a drop-in replacement for attention.}
    \label{fig:hyena_block}
\end{wrapfigure}

Given an input $x \in \mathbb{R}^{L}$ ($L$ denotes sequence length), a ${\sf Hyena}$\footnote{We discuss $D=1$ and order $2$ ${\sf Hyena}$ operators for simplicity.} operator can be defined as:
\begin{equation}\label{eq:linear_attention}
    \begin{aligned}
            (x_1, x_2, v) &\mapsto \sH(x_1, x_2) v \\
            \sH(x_1, x_2) &= {\color{red!60}\sD_{x_2}} {\color{blue!60}\sT_h}{\color{red!60}\sD_{x_1}} 
    \end{aligned}
\end{equation}
where $x_1$, $x_2$, $v$ are projections of the input, and $\sT_h\in\R^{L\times L}$ is the Toeplitz matrix constructed from a learnable long convolution filter produced as the output of a neural network, $(\sT_h)_{ij} = h_{i-j}$. The convolution filter values themselves are obtained through a small neural network $\gamma_\theta$ taking as input the time (position) index and optionally positional encodings, $h_t = \gamma_\theta(t)$, which enable the operator to process very long sequences without growing linearly in the number of parameters. Further, the matrices $\sD_{x_1},\sD_{x_2}\in\R^{L\times L}$ are constructed with $x_1,x_2$ on the diagonals, and evaluated as element-wise gating. The projections are obtained by applying a dense linear layer and short convolution to the input sequence, as shown in Figure \ref{fig:hyena_block}. 

\begin{proposition}A ${\sf Hyena}$ operator
can be evaluated in $\cO(L \log_2 L)$ time. 
\end{proposition}

Efficient evaluation is crucial on settings involving extremely long sequences such as genomics. In the general case where the embedding dimension $D > 1$ and $x\in\R^{L\times D}$, the linear projections $\sW_{x_1}, \sW_{x_2}, \sW_v\in\R^{D\times D}$ are right multiplied to $x$, and $D$ independent ${\sf Hyena}$ operators are then applied to each dimension.

\subsection{Training Long Sequence Models}

\paragraph{Tokenization} The subquadratic cost of ${\sf HyenaDNA}$ in sequence length allows the model to process ultralong sequences directly at the single nucleotide level without the need for frequency-based aggregation tokenizers. This enables fine-grain resolution for both short and long sequences, critical for detecting single nucleotide polymorphisms or mutations and modeling long-range dependencies in gene expression.

We use the natural DNA vocabulary and refer to each nucleotide as a token. The tokens include "A", "G", "C", "T", and "N" (a non-specific nucleotide) and special character tokens for padding, separation, and unknown characters. Tokens are mapped to embedding dimension $D$.

\begin{wrapfigure}[18]{r}{0.45\linewidth}
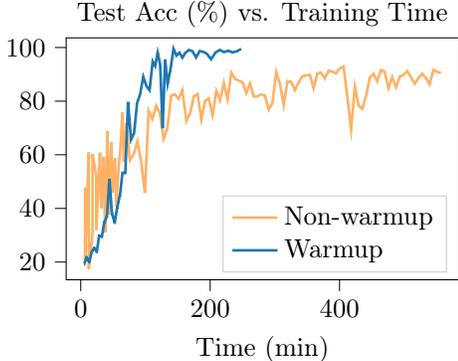

    \centering
    \vspace{-1mm}
    \include{figs/source/warmup}
    \vspace{-10mm}
    \caption{Sequence length warm-up reduces the training time of {$\sf HyenaDNA$} at sequence length 450k by 40\% and boosts accuracy by 7.5 points on species classification.}
    \label{fig:warmup_acc}
\end{wrapfigure}

\paragraph{Sequence length warm-up for ultralong sequences}
\label{warmup}
Directly training on long sequences can affect training stability as the variance in gradient increases \citep{li2022stability}. Training on shorter sequences initially (followed by longer sequences) was used by \citep{press2020shortformer} to train small scale Transformers and reduce training time, while \citep{li2022stability} used sequence length warm-up to address stability on up to 2k tokens.
For ultralong sequences (200k+), we develop a new warm-up schedule that gradually increases the sequence length in stages to improve both stability and decrease training time.

Our sequence length schedule starts at $L_1 = 64$, then doubles the window at each stage while keeping the global batch size constant. By doing so, iterations at each consecutive stage will include more tokens, ensuring the scheduler can also act as a form of batch size warm-up.  
In Fig. \ref{fig:warmup_acc}, we observe sequence length scheduling to be particularly important at sequence lengths greater than $450$k, where at this length training time is reduced by 40\% and improving ultimate accuracy by 7.5\% points for a species classification task described later in section \ref{long-range:species}.

\subsection{Downstream Adaptation}

\paragraph{Tuneable prompting for long-context models}

Prompts have been traditionally used to guide the output of a FM \citep{liu2023pre} by prepending additional context to an input.
Expanding on this approach, \textit{soft} tuneable prompting was introduced to inject \textit{learnable} tokens (as weights) into the input directly \citep{lester2021power} as an alternative to model fine-tuning.

With an extended context length ($L$), we're able to explore new paradigms in adapting FMs after pretraining. Given a downstream task with prompts $x_p\in\R^T$ and corresponding labels $y_p$, we prepend $N \leq L - T$ trainable parameters $\theta$ of dimension $D$ after the embedding step:
\begin{equation}\label{eq:linear_attention}
    \begin{aligned}
    x \leftarrow {\tt concat}[{\tt embed}(x_p), \theta], \quad x\in{\R^{L\times (T + N)}}
    \end{aligned}
\end{equation}

The resulting sequences $x$ are then processed by the model, and $\theta$ is optimized on a loss function involving the input sequence's label $y_p$. Crucially, soft prompting requires utilization of a small subset of prompt and label pairs to optimize $\theta$.

During soft prompting, {$\sf HyenaDNA$} only optimizes the parameters of the prompt in the input sequence while keeping all other model parameters fixed.
Soft prompting thereby provides a flexible and computationally efficient approach to adapting genomic FMs to new downstream tasks.

%% file: figs/source/hyena_operator.tex
\begin{tikzpicture}[baseline=2.5ex]
        \node (u) at (-1, 0.65) {$x$};

        \node (Hvc) [minimum width=0.87cm, fill=green!20, blur shadow={shadow blur steps=5}] at (-0.2,0) {$\sW_{v}$};
        \node (Hv) [minimum width=0.81cm, fill=blue!10, blur shadow={shadow blur steps=5}] at (1,0) {$\sT_{v}$};
        \node (Hkc) [fill=green!20, blur shadow={shadow blur steps=5}] at (-0.2,.65) {$\sW_{x_1}$};
        \node (Hk) [fill=blue!10, blur shadow={shadow blur steps=5}] at (1,.65) {$\sT_{x_1}$};
        \node (Hqc) [fill=green!20, blur shadow={shadow blur steps=5}] at (-0.2,1.3) {$\sW_{x_2}$};
        \node (Hq) [fill=blue!10, blur shadow={shadow blur steps=5}] at (1,1.3) {$\sT_{x_2}$};
        \node (s) [fill=red!20, blur shadow={shadow blur steps=5}] at (2.2,0) {$\sD_{x_1}$};
        \node (H2) [fill=blue!20, blur shadow={shadow blur steps=5}] at (3.2,0) {$\sT_h$};
        \node (s1) [fill=red!20, blur shadow={shadow blur steps=5}] at (4.2,0) {$\sD_{x_2}$};
        \node (y)  at (4.8,0) {};
        \draw[->] (Hvc.east) |- (Hv.west);
        \draw[->] (Hqc.east) |- (Hq.west);
        \draw[->] (Hkc.east) |- (Hk.west);
        \draw[->] (u) |- (Hvc.west); 
        \draw[->] (u) |- (Hqc.west); 
        \draw[->] (u) -- (Hkc.west); 
        \draw[-] (Hk.east)  node [above right] {$x_1$}  -| (s); 
        \draw[-] (Hq.east) node [above right] {$x_2$} -| (s1); 
        \draw[->] (Hv.east) node [above right] {$v$} -- (s);
        \draw[->] (s.east) -- (H2); 
        \draw[->] (H2.east) -- (s1);
\end{tikzpicture}

%% file: figs/source/warmup.tex
\begin{tikzpicture}

\definecolor{darkgray176}{RGB}{176,176,176}
\definecolor{lightgray204}{RGB}{204,204,204}
\definecolor{sandybrown255178102}{RGB}{255,178,102}
\definecolor{steelblue31119180}{RGB}{31,119,180}

\begin{axis}[
height=4.8cm,
width=6.8cm,
legend cell align={left},
legend style={
  fill opacity=0.8,
  draw opacity=1,
  text opacity=1,
  at={(0.97,0.03)},
  anchor=south east,
  draw=lightgray204
},
tick align=outside,
tick pos=left,
title={Test Acc (\%) vs. Training Time},
title style={yshift=-2mm},
x grid style={darkgray176},
xlabel={Time (min)},
xmin=-22.550518365701, xmax=584.217584327857,
xtick style={color=black},
y grid style={darkgray176},
ymin=13.3999996632337, ymax=103.60000051558,
ytick style={color=black}
]
\addplot [semithick, sandybrown255178102, line width=1.0pt]
table {%
6.59160565535227 47.7999985218048
6.60158856312434 20.100000500679
12.3812612891197 60.9000027179718
12.3921871860822 17.4999997019768
18.1674077272415 30.5999994277954
18.1693278034528 60.2999985218048
23.9512791117032 52.3000001907349
23.9526063521703 31.9999992847443
29.7326450149218 60.7999980449677
31.7662822047869 40.0999993085861
35.514240805308 59.2000007629395
37.555056031545 31.0000002384186
41.297183072567 68.8000023365021
43.3372937281926 37.7999991178513
47.0892248590787 64.7000014781952
49.1195834318797 45.8999991416931
52.8718001087507 58.7000012397766
54.9093804915746 40.9000009298325
58.6534579316775 55.5999994277954
60.702220761776 60.3999972343445
64.4357239286105 75.7000029087067
66.4978729685148 57.3000013828278
70.2262508273125 71.7000007629395
76.0080193797747 57.9999983310699
81.7890270034472 60.1999998092651
87.5722349047661 64.9999976158142
93.3551704565684 57.4000000953674
99.1373341242472 45.8000004291534
104.919719163577 76.3999998569489
110.711287784576 73.2999980449677
116.492796579997 78.2999992370605
122.284434998035 75.5999982357025
128.069253162543 66.0000026226044
133.851450296243 69.8000013828278
139.633385888735 81.9999992847443
145.423577642441 82.5999975204468
151.212547834714 80.9000015258789
156.99873200655 84.1000020503998
163.264870194594 72.6000010967255
169.050477413336 75.4999995231628
174.83362184763 81.8000018596649
180.616520019372 82.5999975204468
186.401285175482 73.199999332428
192.184258667628 80.0999999046326
197.966218229135 81.0000002384186
203.747608963648 74.8000025749207
209.531992900372 83.7999999523163
215.317462193966 80.2999973297119
221.100990875562 86.7999970912933
226.889913256963 80.6999981403351
232.670758628845 90.6000018119812
238.463743289312 88.3000016212463
244.245248540242 87.6999974250793
250.029349184036 87.0000004768372
255.812749540806 82.9999983310699
261.597994506359 86.599999666214
267.380674429735 81.6999971866608
273.162376880646 81.8000018596649
278.946463902791 82.4999988079071
284.731702888012 82.0999979972839
290.515061410268 76.8999993801117
296.299721566836 79.4000029563904
302.081646466255 90.7999992370606
307.873624630769 86.5000009536743
313.65950272878 78.7000000476837
319.445935531457 87.5
325.228823387623 87.1999979019165
331.01269865036 84.1000020503998
336.799066877365 87.0000004768372
342.585057322184 89.300000667572
348.369939720631 89.6000027656555
354.154821479321 89.8000001907349
359.937854711215 84.799998998642
365.720780940851 90.2999997138977
371.503806575139 90.7000005245209
377.288018186887 88.4999990463257
383.074277301629 91.3999974727631
388.866238999367 91.0000026226044
394.648067760468 90.2999997138977
400.435368255774 92.2999978065491
406.229357842604 92.9000020027161
412.02355825901 81.5999984741211
417.814952115218 69.4000005722046
423.597046450774 85.3999972343445
429.381627380848 78.3999979496002
435.166223414739 77.1000027656555
440.952150416374 81.9999992847443
446.737242611249 88.7000024318695
452.522134852409 90.200001001358
458.306639508406 87.4000012874603
464.090151838462 91.6000008583069
469.874073286851 90.8999979496002
475.658867561817 84.8999977111816
481.440990694364 88.9999985694885
487.225430611769 85.1999998092651
493.008746091525 89.0999972820282
498.791972923279 89.8000001907349
504.575644612312 91.2000000476837
510.358715113004 88.9999985694885
516.141553286711 86.5000009536743
521.924531010787 88.5999977588654
527.707130169868 90.200001001358
533.494944214821 88.5999977588654
539.280905572573 86.1999988555908
545.066111079852 91.6999995708466
550.853226439158 90.8999979496002
556.637216023604 90.499997138977
};
\addlegendentry{Non-warmup}
\addplot [semithick, steelblue31119180, line width=1.0pt]
table {%
5.02984993855158 19.200000166893
5.38167876799901 19.200000166893
8.9747016509374 21.7999994754791
9.32272619009018 21.7999994754791
12.8672894279162 20.0000002980232
13.2107790907224 20.0000002980232
16.7511275966962 23.7000003457069
17.0928147315979 23.7000003457069
20.6220356106758 25.2999991178513
20.9572526494662 25.2999991178513
24.4926848053932 23.8000005483627
24.8244289755821 23.8000005483627
28.3857317884763 29.899999499321
28.7133575757345 29.899999499321
32.2887474258741 29.3999999761581
32.6129996498426 29.3999999761581
36.2181841174761 34.799998998642
36.5401072541873 34.799998998642
40.1576917092005 37.0000004768372
40.4761846065521 37.0000004768372
44.1689552704493 50.5999982357025
44.4857313752174 50.5999982357025
48.1796761592229 38.1000012159347
48.4980192661285 38.1000012159347
52.1809954007467 34.7000002861023
52.4991869409879 34.7000002861023
56.1831928571065 40.2999997138977
56.5001537958781 40.2999997138977
60.3182606339455 44.9000000953674
60.6332065701485 44.9000000953674
64.4523477395375 53.2999992370605
64.765646870931 53.2999992370605
68.5970644156138 53.2000005245209
68.9092579404513 53.2000005245209
73.041533601284 79.6999990940094
77.7269443273544 65.7000005245209
82.4058220585187 68.0999994277954
87.0829214453697 79.2999982833862
91.7606412649155 82.5999975204468
96.6240003069242 89.0999972820282
101.486994508902 85.6000006198883
106.339746288458 84.1000020503998
111.192398520311 94.8000013828278
115.035912319024 93.0999994277954
118.869651683172 98.1000006198883
122.714982485771 93.6999976634979
126.550714349747 69.9000000953674
130.384698251883 95.5999970436096
134.219629816214 86.5000009536743
138.783791848024 92.00000166893
143.346814521154 99.5000004768372
147.920251095295 97.000002861023
152.483471894264 98.1000006198883
157.048119433721 96.2000012397766
161.612264529864 98.0000019073486
167.149414976438 99.099999666214
172.687703152498 98.6000001430512
178.223675918579 96.2999999523163
183.760323794683 98.6999988555908
189.29651081562 98.4000027179718
194.833902128537 97.6999998092651
201.477796924114 95.5999970436096
208.122018476327 98.1000006198883
214.769088590145 99.099999666214
221.412866326173 98.2999980449676
228.055760594209 98.7999975681305
234.700953555107 98.199999332428
241.346422374249 98.6000001430512
247.990388623873 99.5000004768372
};
\addlegendentry{Warmup}
\end{axis}

\end{tikzpicture}

%% file: hyenadna/4_experiments.tex
\section{Experiments}
In \ref{experiments:pretraining}, we start with pretraining {$\sf HyenaDNA$} on the human reference genome \citep{grch38}. We then evaluate {$\sf HyenaDNA$} on existing short-range (<5k nucleotides) downstream benchmarks in \ref{experiments:short-range} to assess the performance of single nucleotide resolution. In \ref{experiments:icl}, we explore what new capabilities emerge with longer range genomic modeling in the form of in-context learning. Finally, we push the limits of ultralong context performance in \ref{experiments:long-range}.


\subsection{Pretraining on the Human Genome}
\label{experiments:pretraining}

\begin{wrapfigure}[16]{r}{0.50\linewidth}
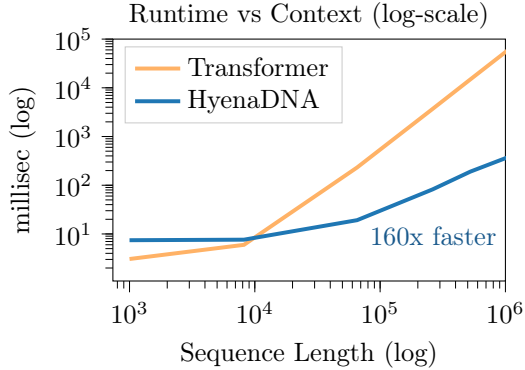

    \vspace{-6mm}
    \centering
    \include{figs/source/runtime}
    \vspace{-11mm}
    \caption{Runtime (forward \& backward pass) for Transformer and {$\sf HyenaDNA$}: 2 layers, width=128, gradient checkpointing, batch size=1, A100 80GB. At 1M tokens {$\sf HyenaDNA$} is \textbf{160x faster} than Transformer.}
    \label{fig:runtime_experiment}
\end{wrapfigure}

We pretrain {$\sf HyenaDNA$} on the human reference genome \citep{grch38} using next nucleotide (token) prediction. Starting with a stack of decoder-only Transformer blocks, we swap attention for the Hyena operator, and compare against a baseline Transformer (GPT) with Flash Attention \citep{dao2022flashattention}. We add gradient checkpointing to {$\sf HyenaDNA$} to decrease the memory footprint by 3x on longer sequences ( > 160k). We then scale {$\sf HyenaDNA$} along dimensions of model depth (2 to 8 layers), width (128 to 256 dimensions), and sequence length (1024 to 1M). At sequence length 1M, {$\sf HyenaDNA$} is 160x faster than its Transformer counterpart as shown in Fig. \ref{fig:runtime_experiment}.

As shown in Fig. \ref{fig:ppl_seqlen}, we observe that as context length increases, perplexity improves during pretraining. However, this improvement comes at the expense of more training time and tokens. For models too shallow to effectively process longer context, perplexity can begin to degrade (increase), observing inflection points with longer sequences. In this way, increasing context can serve as a novel regularization dimension. For genomic pretraining, we provide the following guidelines. 1. In optimizing for faster training time, shorter context enable lower perplexity to be reached faster. 2. In optimizing for best overall perplexity, longer context allows for lower perplexity at the cost of training on more tokens. See \ref{appendix:pretraining-details} for experiment details.


\subsection{Single Nucleotide Resolution}
\label{experiments:short-range}

Our first downstream tasks use short-range genomic sequences (<5k) aimed at evaluating single nucleotide resolution performance on sequence-level classification using standard fine-tuning.

\input{tables/source/genomic_benchmarks}

\paragraph{GenomicBenchmarks}
We start with the newly released GenomicBenchmarks \citep{gresova2022genomic}, which is comprised of 8 regulatory element classification datasets with sequence lengths of 200-500, and one up to 4,776. The original baseline model uses a short-range CNN. We fine-tune the pretrained Transformer (GPT) and {$\sf HyenaDNA$} from \ref{experiments:pretraining}, both having single nucleotide resolution, as well as the DNABERT model \citep{ji2021dnabert}. {$\sf HyenaDNA$} sets a new SotA on 7 of 8 datasets and by up to 20\% points on the human enhancer identification task, as shown in Tab. \ref{tab:genomic-benchmark}. See \ref{appendix:short-range-details} for additional experiment details and ablations.

\paragraph{Nucleotide Transformer} 
Next, we benchmark against 18 datasets from the Nucleotide Transformer (NT) \citep{dallatorre2023nucleotide}, which includes predicting regulatory elements for enhancers, promoters, epigenetic marks, and splice sites from DNA sequences of length 200-600 nucleotides. We compare against 3 NT base models, which were pretrained using masked language modeling (BERT) and then fine-tuned. The NT models ranged from 500M to 2.5B parameters, and pretrained on up to 3202 genomes. All NT models use 6-mer sequences of 1000 tokens long. For {$\sf HyenaDNA$}, we attach a linear decoder head and fine-tune a pretrained model, surpassing SotA on 12 of 18 datasets using a model with orders of magnitude less parameters and pretraining data, shown in Tab. \ref{tab:nuctran}. See \ref{appendix:short-range-details} for additional experiment details and ablations.


\subsection{In-context Learning for Genomic Sequences}
\label{experiments:icl}

Compared to natural language FMs, which have shown strong success with in-context learning, {$\sf HyenaDNA$}'s vocabulary is very small. DNA sequences are also less diverse in structure, e.g. there's no concept of labels or descriptions that follow a DNA sequence.
This makes it challenging to perform "pure" in-context learning (relying only on inference), since new concepts such as classification labels would require new symbols.
\input{tables/source/nuctrans}
To overcome this limitation and explore the potential for in-context learning in genomics, we make use of two variants of in-context learning: soft prompting and instruction fine-tuning. Each involve a brief tuning phase to introduce the concept of classification using only the existing vocabulary.

\paragraph{Procedure}
 
In both variants, we use the GenomicBenchmarks in \ref{experiments:short-range}, and a {$\sf HyenaDNA$} model pretrained on sequence length 160k from \ref{experiments:pretraining}. 

In the first experiment, we evaluate a soft prompting approach by prepending a sequence of soft tuneable tokens (2 to 32k) directly in the input sequences. We include a brief tuning phase ($<$ 20 epochs), updating the soft tokens only, to provide {$\sf HyenaDNA$} with the ability to indicate the target classes. To denote classes, we repurpose {$\sf HyenaDNA$}'s fixed vocabulary: for binary classification, for example, we indicate the two classes with the letters "A" and "N".

In the second experiment, we evaluate a few-shot learning approach to in-context learning \citep{brown2020language} by prepending, consecutively, $k$ (2 to 32) demonstrations of each class and its sequence into the prompt.
As before, we encode class labels by the use of individual letters of {$\sf HyenaDNA$}'s existing vocabulary.
We additionally perform a brief instruction-tuning period \citep{wei2021finetuned} for each dataset to familiarize {$\sf HyenaDNA$} with this task structure by tuning the pretrained model on a small subset of the dataset.

\begin{figure}[!h]
    \centering
    \includegraphics[width=\linewidth]{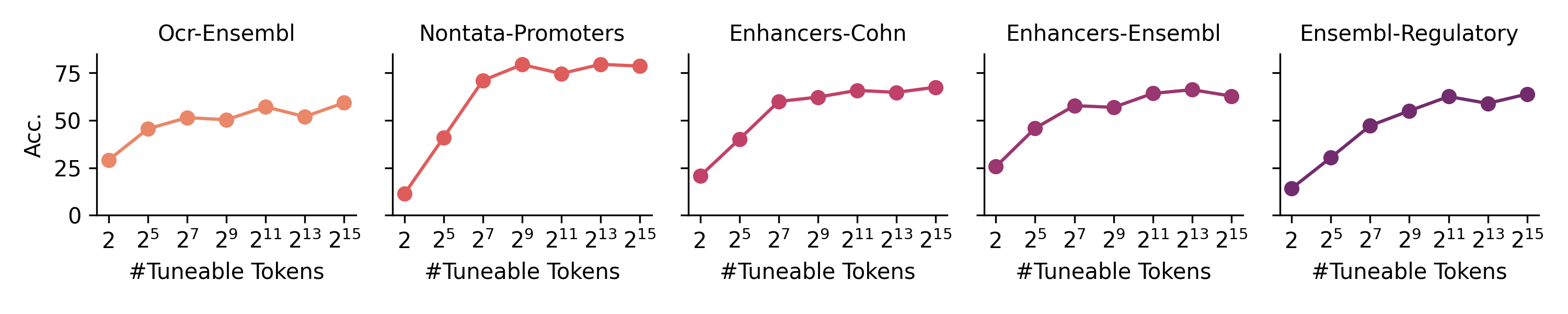}
    \vspace{-6mm}
    \caption{{\bf Filling long-context with soft tuneable tokens.} {$\sf HyenaDNA$} is able to learn new tasks in-context when adding a sequence of tuneable tokens to the input sequences. Longer sequences of tuneable tokens lead to better performance.}
    \label{fig:soft-prompting}
\end{figure}

\paragraph{Results} In Fig.\ \ref{fig:soft-prompting}, {$\sf HyenaDNA$}'s performance on novel tasks improves as more tuneable tokens are added into the input sequences, and saturates close to baseline performance (Tab. ~\ref{tab:genomic-benchmark}; with the exception of the Human Regulatory dataset).
By contrast, we find that increasing $k$-shot demonstrations to the input does not necessarily improve performance. A higher number of tuning samples is needed before $k$-shot demonstrations start to boost accuracy as shown in Tab. \ref{fig:fewshot-prompting}. See \ref{appendix:icl-details} for experiment details.


\subsection{Ultralong-Range Genomics}
\label{experiments:long-range}

\input{tables/source/chromatin}

In our final experimental section, we focus on pushing the limits of using long context effectively in genomics. In \ref{long-range:chromatin}, we tackle a challenging 919 binary multi-task against a sparse-attention baseline. In \ref{long-range:biotype} we analyze the learned embeddings {$\sf HyenaDNA$} and its use in clustering long sequences by functional annotation, and in \ref{long-range:species} we showcase a novel ultralong-range species classification task.


\subsubsection{Chromatin Profile Prediction}
\label{long-range:chromatin}

The prediction of chromatin profiles and epigenetic markers from DNA sequences is an important and challenging task to quantify the functional effects of non-coding variants. These variants include single nucleotide changes in DNA that can affect the downstream expression of genes \citep{zaina2010genetics}. The DeepSEA dataset \citep{zhou2015predicting} is compiled from 919 chromatin features including transcription factor (TF) binding profiles, DNase I-hypersensitive sites (DHS) and histone mark (HM) profiles. For a given sequence, the task is to jointly predict 919 labels corresponding to the chromatin profile (similar to peak detection) of a central region of the sequence, indicating the presence of such functional effects. The input also includes flanking regions that provide broader contextual information needed to incorporate long-range interactions. We fine-tune our pretrained {$\sf HyenaDNA$} models from \ref{experiments:pretraining} and perform competitively against a DeepSea CNN and the SotA sparse attention BigBird \citep{zaheer2020big} baselines using 5-30$\times$ fewer parameters. See \ref{appendix:chromatin-details} for experiment details.


\subsubsection{Biotype Embeddings}
\label{long-range:biotype}


\begin{figure}[h]
    \centering
    \includegraphics[width=1.0\linewidth]{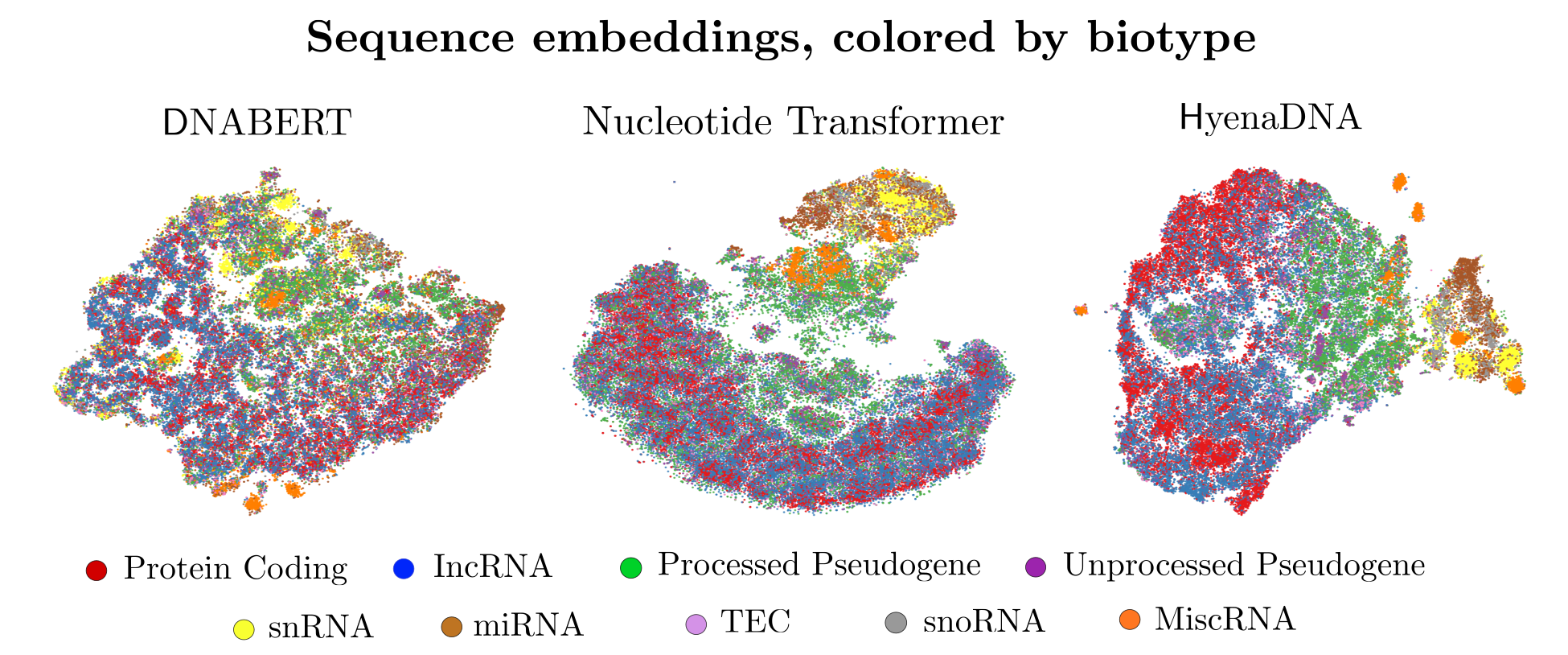}
    \vspace{2mm}
    \caption{\textbf{Embedding visualisation.} t-SNE of the embeddings generated by DNABERT, Nucleotide Transformer and {$\sf HyenaDNA$} coloured by Ensembl biotype annotations.}
    \vspace{3mm}
\label{fig:t-sne_embeddings}
\end{figure}

\noindent Next, we analyze the pretrained embeddings from {$\sf HyenaDNA$} and compare them with DNABERT \citep{ji2021dnabert} and the Nucleotide Transformer \citep{dallatorre2023nucleotide}. 
We encode sequences of human genes corresponding to different biological function annotations obtained from the Ensembl dataset known as biotypes \citep{cunningham2022ensembl}. In cases where the length of the input exceeds the context window of the encoder, the sequence is chunked (by the max length of the encoder) and averaged.

\input{tables/source/biotype}

We fit the embeddings using an XGBoost \citep{chen2016xgboost} classifier on the 10 most frequent biotypes, and apply t-SNE \citep{van2008visualizing} for visualization. As shown in \ref{fig:t-sne_embeddings}, distinct clusterings emerge visually, while quantitatively, {$\sf HyenaDNA$} produces the highest F1 score in biotype classification (with a much smaller model), indicating that during pretraining, {$\sf HyenaDNA$} learns informative features related to biological function.


\subsubsection{Species Classification}
\label{long-range:species}

The majority of the genome is conserved across species -- humans and non-human primates, for example, have <10\% sequence divergence \citep{rogers2014comparative}, making them difficult to discriminate. This allows us to to design an ultralong-range sequence modeling task to test whether a model can determine the source species of a random genetic sequence. To train, we randomly sample DNA sequences from 5 different species, and fine-tune pretrained {$\sf HyenaDNA$} and Transformer models from \ref{experiments:pretraining} to predict the species label.
We observe in Tab. \ref{tab:species_classification} that both models struggle on shorter sequences of length $1024$, but performance improves with longer contexts as the distinct mutational profile of each species becomes more evident. {$\sf HyenaDNA$} effectively solves the task by using a context length of $450$k to $1$ million, where Transformer cannot due to infeasible training time limitations. See \ref{appendix:species-details} for experiment details.

\input{tables/source/species}

%% file: figs/source/runtime.tex
\begin{tikzpicture}

\definecolor{darkgray176}{RGB}{176,176,176}
\definecolor{lightgray204}{RGB}{204,204,204}
\definecolor{sandybrown255178102}{RGB}{255,178,102}
\definecolor{steelblue31119180}{RGB}{31,119,180}

\begin{axis}[
height=4.8cm,
width=6.8cm,
legend cell align={left},
legend style={
  fill opacity=0.8,
  draw opacity=1,
  text opacity=1,
  at={(0.03,0.97)},
  anchor=north west,
  draw=lightgray204
},
log basis x={10},
tick align=outside,
tick pos=left,
title={Runtime vs Context (log-scale)},
title style={yshift=-2mm},
x grid style={darkgray176},
xlabel={Sequence Length (log)},
xmin=736.782622705323, xmax=1000000,
xmode=log,
xtick style={color=black},
y grid style={darkgray176},
ylabel={millisec (log)},
ytick style={color=black},
ymode=log,
ymax=100000,
ytick={ 1, 10, 100, 1000, 10000, 100000, 1000000},
]
\addplot [semithick, sandybrown255178102, line width=1.5pt]
table {%
    1000 3.06
    8192 5.98
    65536 231.7
    262144 3680
    524288 14840
    1048576 59490
};
\addlegendentry{Transformer}
\addplot [semithick, steelblue31119180, line width=1.5pt]
table {%
    1000 7.44
    8192 7.63
    65536 19.19
    262144 82.11
    524288 190.06
    1048576 376.4
};
\addlegendentry{HyenaDNA}

\node [anchor=west, text=darkersteelblue] at (axis cs:70000, 9) {160x faster};

\end{axis}

\end{tikzpicture}

%% file: tables/source/genomic_benchmarks.tex
\begin{wraptable}[14]{r}{0.65\textwidth}  
      \vspace{-8mm}
      \small
      \caption{{\bf GenomicBenchmarks}
        Top-1 accuracy (\%) for pretrained {$\sf HyenaDNA$}, DNABERT and Transformer (GPT from \ref{experiments:pretraining}), and the previous SotA baseline CNN (scratch).
      }
      \label{tab:genomic-benchmark}
    \vspace{4mm}
    \centering
    \setlength{\tabcolsep}{4pt} 
        \begin{tabular}{lcccc}
            \toprule
            \textsc{Dataset} & \textsc{CNN} & \textsc{DNABERT} & \textsc{GPT} & \textsc{HyenaDNA} \\
            \midrule
            Mouse Enhancers & 69.0 & 66.9 & 80.1 & \textbf{85.1} \\
            Coding vs Intergenomic & 87.6 & \textbf{92.5 }& 88.8 & 91.3 \\
            Human vs Worm & 93.0 & 96.5 & 95.6 & \textbf{96.6} \\
            Human Enhancers Cohn & 69.5 & 74.0 & 70.5 & \textbf{74.2} \\
            Human Enhancers Ensembl & 68.9 & 85.7 & 83.5 & \textbf{89.2} \\
            Human Regulatory & 93.3 & 88.1 & 91.5 & \textbf{93.8} \\
            Human Nontata Promoters & 84.6 & 85.6 & 87.7 & \textbf{96.6} \\
            Human OCR Ensembl & 68.0 & 75.1 & 73.0 & \textbf{80.9} \\
            \bottomrule
        \end{tabular}
\end{wraptable}


%% file: tables/source/nuctrans.tex
\begin{wraptable}[26]{r}{0.58\textwidth} 
\setlength{\tabcolsep}{5pt}
    \vspace{-5mm}
    \small
    \caption{{\bf Nucleotide Transformer (NT) Benchmarks} 
      The Matthews correlation coefficient (MCC) is used as the performance metric for the enhancer and epigenetic marks dataset, and the F1-score is used for the promoter and splice site dataset.}
    \label{tab:nuctran}
    \vspace{5mm}
    \centering
        \begin{tabular}{lcccc}
            \toprule
            \textsc{Model} 
            & \textsc{NT} & \textsc{NT} & \textsc{NT} & {$\sf HyenaDNA$} \\
            \textsc{Params} & 500M & 2.5B & 2.5B & 1.6M \\
            \textsc{\# of Genomes} & 1 & 3,202 & 850 & 1 \\
            \midrule
            Enhancer
            & 53.5 & 59.3 & 58.0 & \textbf{62.6} \\
            Enhancer types
            & 48.5 & 50.0 & 47.4 & \textbf{55.7} \\
            H3
            & 73.7 & 77.6 & 81.4 & \textbf{81.7} \\
            H3K4me1
            & 35.8 & 44.5 & 55.9 & \textbf{57.1} \\
            H3K4me2
            & 28.1 & 30.0 & 32.6 & \textbf{53.9} \\
            H3K4me3
            & 26.3 & 28.1 & 42.1 & \textbf{61.2} \\
            H3K9ac
            & 46.2 & 50.8 & 57.5 & \textbf{65.1} \\
            H3K14ac
            &  37.7 & 47.1 & 55.0 & \textbf{66.3} \\
            H3K36me3
            & 46.7 & 53.3 & 63.2 & \textbf{65.3} \\
            H3K79me3
            & 57.7 & 59.2 & 64.2 & \textbf{71.6} \\
            H4
            & 76.2 & 78.9 & \textbf{82.2} & 79.6 \\
            H4ac
            & 34.4 & 42.3 & 50.1 & \textbf{63.7} \\
            Promoter all
            &  95.4 & 96.6 & \textbf{97.4} & 96.5 \\
            Promoter non-TATA
            & 95.6 & 96.9 &\textbf{97.7} & 96.6 \\
            Promoter TATA
            & 94.8 & 95.8 & 96.4 & \textbf{96.7} \\
            Splice acceptor
            & 96.5 & 98.5 & \textbf{99.0} & 96.6 \\
            Splice donor
            & 97.2 & 98.2 & \textbf{98.4} & 97.3 \\
            Splice all 
            & 97.2 & 97.8 & \textbf{98.3} & 97.9 \\
            \bottomrule
        \end{tabular}
\end{wraptable}

%% file: tables/source/chromatin.tex
\begin{wraptable}[12]{h}{0.53\textwidth}
    \vspace{-8mm}
    \small
    \caption{{\bf Chromatin profile prediction}
        Median AUROC computed over three categories: Transcription factor binding profiles (TF), DNase I-hypersensitive sites (DHS) and histone marks (HM). 
    }
    \vspace{3mm}
    \centering
    \begin{tabular}{crcccc}
        \toprule
        \multirow{2}{*}{\sc{Model}}  & \multirow{2}{*}{\sc{Params}} & \multirow{2}{*}{\sc{Len}} & \multicolumn{3}{c}{\sc{AUROC}}\\
        & & & \sc{TF} & \sc{DHS} & \sc{HM} \\\midrule
        DeepSEA & 40 M & 1k &  95.8 & 92.3 & 85.6 \\
        BigBird & 110 M & 8k &  96.1 & 92.1 & 88.7 \\\midrule
        \multirow{2}{*}{{$\sf HyenaDNA$}} 
        & 7 M & 1k & \textbf{96.4} & \textbf{93.0} & 86.3 \\
        & 3.5 M & 8k  & 95.5 & 91.7 & \textbf{89.3} \\
        \bottomrule
        \end{tabular}
        \label{tab:chromatin_profile_aucs}
\end{wraptable}

%% file: tables/source/biotype.tex
\begin{wraptable}[9]{h}{0.42\textwidth}
    \vspace{-4mm}
    \small
    \caption{\textbf{Embedding quality} Weighted F1 classification score on $10$ biotypes.}
    \vspace{1mm}
    \centering
    \begin{tabular}{crccc}
    \toprule
       \sc{Model} & \sc{Params} & \sc{Len} & \sc{F1}  \\
       \midrule 
       DNABERT & $110$ M & 512 & 64.6  \\
       NT & $500$ M & 6k & 66.5  \\
       \midrule
       {$\sf HyenaDNA$} & 7 M  & 160k & $\mathbf{72.0}$\\
       \bottomrule
    \end{tabular}
    \label{tab:embedding_downstream}
\end{wraptable}

%% file: tables/source/species.tex
\begin{wraptable}[18]{r}{0.38\linewidth}
    \vspace{-8mm}
    \small
    \caption{{\bf Species classification}
        Top-1 accuracy (\%) for 5-way classification (human, lemur, mouse, pig, hippo). The \xmark~ symbol indicates infeasible training time.
    }
    \vspace{2mm}
    \centering
    \begin{tabular}{lcc}
        \toprule
        \sc{Model} & \sc{Len} & \sc{Acc} \\\midrule
        Transformer & 1k & 55.4 \\
        ${\sf HyenaDNA}$ & 1k & 61.1 \\
        \midrule
        Transformer & 32k & 88.9 \\
        ${\sf HyenaDNA}$ & 32k & 93.4 \\
        \midrule
        Transformer & 250k & \xmark \\
        ${\sf HyenaDNA}$ & 250k & 97.9\\
        \midrule
        Transformer & 450k & \xmark \\
        ${\sf HyenaDNA}$ & 450k & 99.4\\
        \midrule
        Transformer & 1M & \xmark \\
        ${\sf HyenaDNA}$ & 1M & \textbf{99.5}\\
        \bottomrule
        \end{tabular}
        \label{tab:species_classification}
\end{wraptable}

%% file: hyenadna/5_conclusion.tex
\section{Conclusion}

\paragraph{Summary} 

We presented {$\sf HyenaDNA$}, a genomic foundation model pretrained on the human reference genome with context lengths up to 1 million tokens at single nucleotide resolution - an up to 500x increase over previous genomic FMs using dense-attention. {$\sf HyenaDNA$} is able to learn generalizable features that can then be fine-tuned for tasks including identifying regulatory elements and on a 919-way chromatin profile prediction task. We also explored the first use of in-context learning in genomics to enable simpler adaptation to downstream tasks without any updates to pretrained weights. 

\paragraph{Limitations and Future Work}

While demonstrating competitive results and introducing novel capabilities, it is worth noting that {$\sf HyenaDNA$} was pretrained on only one human reference genome. Incorporating genomes of multiple humans and species could increase generalizability in learned features and reduce bias. Furthermore, our current focus in this study was exclusively on DNA sequences. Extending our framework to incorporate other biological or chemical sequences, such as proteins and drug molecules, has the potential to unlock multi-modal capabilities similar to those observed in natural language and vision FMs \citep{radford2021learning, ramesh2021zero, yu2022coca}. 

With respect to model size, {$\sf HyenaDNA$} is significantly smaller than previous genomic FMs and was pretrained using up to 8 Nvidia A100 (80GB) GPUs. We expect increasing model size, and compute, may lead to additional long-range capabilities. Notably, with model parallelism, it becomes feasible to extend the context length by orders of magnitude beyond this current work, and leave that open to future research. 

Furthermore, beyond discriminative applications, the use of long context models in generative tasks unlocks exciting prospects for the design of synthetic regulatory elements, genes and protein complexes. In conclusion, the continued advancements of long-range sequence models with single nucleotide resolution hold great promise in driving innovation in genomic research and unraveling the complexities of biological systems.

%% file: hyenadna/6_ack.tex
\section*{Acknowledgments}

We would like to thank Guatam Machiraju, Elliott Epstein, Archis Joglekar, Jared Dunnmon, Nazim Bouatta and Anshul Kundaje for helpful discussion and feedback on earlier drafts, and Together for providing the compute used to train models in this paper. We gratefully acknowledge the support of NIH under No. U54EB020405 (Mobilize), NSF under Nos. CCF1763315 (Beyond Sparsity), CCF1563078 (Volume to Velocity), and 1937301 (RTML); US DEVCOM ARL under No. W911NF-21-2-0251 (Interactive Human-AI Teaming); ONR under No. N000141712266 (Unifying Weak Supervision); ONR N00014-20-1-2480: Understanding and Applying Non-Euclidean Geometry in Machine Learning; N000142012275 (NEPTUNE); NXP, Xilinx, LETI-CEA, Intel, IBM, Microsoft, NEC, Toshiba, TSMC, ARM, Hitachi, BASF, Accenture, Ericsson, Qualcomm, Analog Devices, Google Cloud, Salesforce, Total, the HAI-GCP Cloud Credits for Research program,  the Stanford Data Science Initiative (SDSI), Department of Defense (DoD) through the National Defense Science and Engineering Graduate Fellowship (NDSEG) Program, and members of the Stanford DAWN project: Facebook, Google, and VMWare. This work is supported by NSF (1651565), AFOSR (FA95501910024), ARO (W911NF-21-1-0125), ONR, DOE (DE-SC0022222), CZ Biohub, and Sloan Fellowship. The U.S. Government is authorized to reproduce and distribute reprints for Governmental purposes notwithstanding any copyright notation thereon. Any opinions, findings, and conclusions or recommendations expressed in this material are those of the authors and do not necessarily reflect the views, policies, or endorsements, either expressed or implied, of NIH, ONR, or the U.S. Government. 

%% file: hyenadna/appendix/experiments.tex
\section{Appendix: Experimental Details}
\label{appendix:experimental-details}

In the following sections we provide further details for each experiment. Across all experiments, we use Pytorch and Pytorch Lightning. We train on a mix of Nvidia GPUs with A100s, V100s, and T4s. Unless otherwise stated, we use a cross entropy loss for our objective. Our repository is made public here: \href{URL}{https://github.com/HazyResearch/hyena-dna}.

\input{hyenadna/appendix/experiment_subsections/pretraining}

\input{hyenadna/appendix/experiment_subsections/short_range}

\input{hyenadna/appendix/experiment_subsections/icl}

\input{hyenadna/appendix/experiment_subsections/chromatin}

\input{hyenadna/appendix/experiment_subsections/embeddings}

\input{hyenadna/appendix/experiment_subsections/species}

%% file: hyenadna/appendix/experiment_subsections/pretraining.tex
\subsection{Pretraining Details}
\label{appendix:pretraining-details}

\begin{table}[h]
    \small  
    \centering
    \caption{Hyperparameter settings for {$\sf HyenaDNA$} pretraining (select models).}  
    \begin{tabular}{lccccc}
        \toprule
        Layers & 2 & 2 & 4 & 4 & 8 \\
        Width & 128 & 256 & 128 & 256 & 256 \\
        Params (M) & 0.44 & 1.6 & 0.87 & 3.3 & 6.6 \\
        Max seq. len. & 64k & 64k & 64k & 64k & 1M \\
        \midrule
        Optimizer & \multicolumn{5}{c}{AdamW} \\
        Optimizer momentum & \multicolumn{5}{c}{$\beta_1$, $\beta_2$ = 0.9, 0.999} \\
        Learning rate & \multicolumn{5}{c}{1.5 - 6e-4}\\
        LR Scheduler & \multicolumn{5}{c}{Cosine decay} \\
        Batch size & \multicolumn{5}{c}{64 - 256} \\
        Global steps & \multicolumn{5}{c}{10 - 20k} \\
        Weight decay (model) & \multicolumn{5}{c}{0.1}\\
        Weight decay ({$\sf Hyena$} layers) & \multicolumn{5}{c}{0}\\
        Embed dropout & \multicolumn{5}{c}{0.1}\\
        Residual dropout & \multicolumn{5}{c}{0}\\
        \bottomrule
    \end{tabular}
    \label{tab:pretraining_hyper}
\end{table}

\paragraph{Data} For pretraining, we use a single human reference genome \citep{grch38}, and leverage the training and validation intervals (start and end) from \citep{avsec2021effective}. During training, we sample an interval and obtain a sequence of length $L$ by adjusting the intervals on both ends. For the test set, we use chromosomes 14 and X, exclusively, and sample non-overlapping sequences of length $L$.

\paragraph{Model} We design a suite of parameter efficient architectures with depths between 2 and 8 layers, Hyena blocks of Order-N = 2, and width 128 to 256. The MLP expansion factor (reverse bottleneck) is 4x the width. See Fig. \ref{fig:hyena-arch} for the block architecture of {$\sf HyenaDNA$}. The parameter counts range from 400k to 6.6M, trained on sequence lengths between 1,024 and 1M. Tab. \ref{tab:pretraining_hyper} highlights a representative subset of the models we trained. Note: we use different pretrained model sizes depending on the downstream task to prevent overfitting. When selecting which pretrained model to use for a downstream task, we found that a pretrained sequence length of 2 to 4x the downstream max sequence length results in the best performance. 

\paragraph{Training} We pretrain each model for 10-20k global steps. For models trained on longer sequences, this translates to more tokens being used, as each sample contains more tokens. For example, the largest model with context length 1M was trained on 2T tokens over 4 weeks. We adjust the "accumulate\_grad\_batches" argument in Pytorch Lightning to keep the global batch size consistent across models and sequence lengths. See Tab. \ref{tab:pretraining_hyper} for hyperparameter details.

\paragraph{Training efficiency}

We compare pretraining compute resources and GPU-hours to reach competitive performance on the short-range tasks for several baselines and {$\sf HyenaDNA$} models, shown in Tab. \ref{tab:gpu-comparison}.

\begin{table}[h]
    \caption{Pretraining GPU \& runtime comparison for short-range models.}
    \label{tab:gpu-comparison}
    \centering
    \begin{tabular}{lcccc} \toprule
    {} & \sc{DNABERT} & \sc{Nucleotide Transformer} & {$\sf HyenaDNA$} & {$\sf HyenaDNA$} \\ \midrule
    Params & 110M & 2.5B & 436K & 1.6M \\
    GPUs & 8-2080 TI & 128-A100-80GB & 1-A100-40GB & 1-A100-40GB \\
    Wall clock	& 25 days & 28 days & 80 mins & 80 mins \\
    GPU-hrs	& 12,000 & 215,000 & 1.3 & 1.3 \\
    \bottomrule 
    \end{tabular} 
\end{table}

%% file: hyenadna/appendix/experiment_subsections/short_range.tex
\subsection{Short-Range Genomics Details}
\label{appendix:short-range-details}

\subsubsection{GenomicBenchmarks experiment}

\paragraph{Data}

The GenomicBenchmarks \citep{gresova2022genomic} includes 8 datasets designed for sequence-level classification tasks that involve predicting regulatory elements, along with one binary species task. The benchmarks provided for the baseline model include two sets of results: one obtained with Pytorch and the other with TensorFlow. Since our code base is implemented in Pytorch, we compare our results with the Pytorch-based benchmarks.

\paragraph{Model}

Our backbone is a pretrained 2 layer {$\sf HyenaDNA$} model with width 128, trained on sequence length 1024. We pool along the sequence dimension to obtain a classification token, and attach a simple linear decoder head. The baseline CNN, as described by \citep{gresova2022genomic}, uses uses an embedding layer, 3 convolutional layers with number of filters: 16, 8, and 4. It uses batch norm and max pooling after each convolutional layer, followed by 2 dense layers. It is trained for 10 epochs with batch size 64. The mode sizes range from 120k to 520k, depending on sequence length chosen.

\begin{table}[h]
    \small  
    \centering
    \caption{GenomicBenchmarks hyperparameters for {$\sf HyenaDNA$} and the baseline Transformer (GPT from \ref{experiments:pretraining}), which uses FlashAttention \citep{dao2022flashattention}.}  
    \begin{tabular}{lcc}
        \toprule
         & \sc{Transformer} & {$\sf HyenaDNA$} \\
        \midrule
        Layers & 2 & 2 \\
        Width & 128 & 128 \\
        Parameters & 529k & 436k \\
        Learning rate & 1-$6e^{-4}$ & 1-6$e^{-4}$ \\
        Weight decay (model) & 0-0.2 & 0-0.2 \\
        Weight decay ({$\sf Hyena$} layers) & - & 0 \\
        Embed dropout & 0-0.2 & 0.0-0.3 \\ 
        Resid dropout & 0-0.2 & 0-0.3 \\
        Num heads & 8 & - \\
        \midrule
        Optimizer & \multicolumn{2}{c}{AdamW} \\
        Optimizer momentum & \multicolumn{2}{c}{$\beta_1$, $\beta_2$ = 0.9, 0.999} \\
        LR scheduler & \multicolumn{2}{c}{Cosine decay} \\
        Batch size & \multicolumn{2}{c}{128-1024} \\
        Training epoch & \multicolumn{2}{c}{100} \\
        Reverse complement aug. & \multicolumn{2}{c}{true/false} \\
        Sequence lengths & \multicolumn{2}{c}{200-4800} \\
        \bottomrule
    \end{tabular}
    \label{tab:appendix-genomic-benchmark}
\end{table}

\paragraph{Training}

The primary hyperparameters we sweep across include: learning rate, global batch size, dropout, weight decay, and a reverse complement augmentation. See Tab. \ref{tab:appendix-genomic-benchmark} for ranges of hyperparamters used.

\subsubsection{Ablations on the GenomicBenchmarks}

To better understand how specific design choices in the {$\sf HyenaDNA$} model effect performance, we perform a series of ablation experiments on the GenomicBenchmarks.

\begin{table}[h]
      \small
      \caption{{\bf GenomicBenchmarks Top-1 accuracy (\%)} GPT is the causal Transformer from \ref{experiments:pretraining}, $\sf HyenaDNA$ k-mer uses a 6-mer tokenizer, and $\sf HyenaDNA$ bidirection is a bidirectional version of the Hyena operator.
      }
      \label{tab:genomic-benchmark-ablations}
    \vspace{4mm}
    \centering
        \begin{tabular}{lccccccc}
          \toprule
              \multirow{2}{*}{\sc{Model}} & \multirow{2}{*}{\sc{GPT}} & \multirow{2}{*}{\sc{GPT}} & \multirow{2}{*}{$\sf HyenaDNA$} & \multirow{2}{*}{$\sf HyenaDNA$} & {$\sf HyenaDNA$} & {$\sf HyenaDNA$} & \multirow{2}{*}{\sc{DNABERT}} \\
              & & & & & k-mer & bidirection \\
              \midrule
              Pretrained & no & yes & no & yes & no & no & yes \\
              \midrule
              Mouse Enhancers & 79.3 & 79.3 & 84.7 & \textbf{85.1} & 81.8 & 80.6 & 66.9 \\
              Coding vs Intergenomic & 89.3 & 91.2 & 90.9 & 91.3 & 86.7 & 90.3 & \textbf{92.5} \\
              Human vs Worm & 94.8 & \textbf{96.6} & 96.4 & \textbf{96.6 }& 92.9 & 95.9 & 96.5 \\
              Human Enhancers Cohn & 67.7 & 72.9 & 72.9 & \textbf{74.2} & 69.8 & 72.1 & 74.0 \\
              Human Enhancers Ensembl & 79.0 & 88.3 & 85.7 & \textbf{89.2} & 88.0 & 85.9 & 85.7 \\
              Human Regulatory & 90.2 & 91.8 & 90.4 & \textbf{93.8} & 90.2 & 89.1 & 88.1 \\
              Human Nontata Promoters & 85.2 & 90.1 & 93.3 & \textbf{96.6 }& 83.5 & 88.5 & 85.6 \\
              Human OCR Ensembl & 68.3 & 79.9 & 78.8 & \textbf{80.9} & 70.2 & 75.3 & 75.1 \\
            \bottomrule
        \end{tabular}
\end{table}

\paragraph{Pretraining:} We train {$\sf HyenaDNA$} from scratch and compare with the pretrained version. The pretrained models provide mild to moderate gains - likely due to the benchmarks being near saturation already.

\paragraph{Tokenization:} We train {$\sf HyenaDNA$} using a k-mer  tokenizer (k=6) to isolate the effect of the single nucleotide tokenizer. The k-mer tokenizer drops performance significantly across on a majority of the datasets (by as much as 10 accuracy points), while boosting one dataset (Human Enhancer Ensembl). Therefore, the single nucleotide tokenization appears to be a significant component of the {$\sf HyenaDNA$} model.

\paragraph{Bidirectional:} To ablate the impact of using a causal model, we implemented a bidirectional version of {$\sf HyenaDNA$} and trained from scratch on the GenomicBenchmarks (i.e. without masked language model pretraining). The bidirectional version degraded performance on 7 of 8 datasets compared to the standard causal {$\sf HyenaDNA$} (also from scratch), on average by 3.8 accuracy points. 

The bidirectional {$\sf HyenaDNA$} was implemented by using a circular FFT convolution. This involved manipulating the padding on the input sequence before performing the FFT convolution. Previously, we zero padded the input on the right side by length $L$ (the sequence length). For bidirectionality, we pad by $1/2$ $L$ on the left and right side of the input, effectively providing a bidirectional receptive field (due to the circular convolution). This is one of many possible ways to implement a bidirectional version of Hyena.

\subsubsection{Downstream prediction tasks for Nucleotide Transformer benchmark}

Following the Nucleotide Transformer \citep{dallatorre2023nucleotide}, we collected datasets from four different sources \citep{geng2022,pham2005,oubounyt2019deepromoter,scalzitti2021spliceator}. 

\paragraph{Promoter}  The promoter dataset included TATA-box-containing and TATA-box-lacking promoters. Tasks involved predicting promoters with a TATA-box, identifying promoters lacking a TATA-box, and distinguishing between both promoter categories and non-promoter sequences. 
The promoter datasets were obtained from the Eukaryotic Promoter Database (EPDnew)\footnote{https://epd.epfl.ch//index.php} for human and mouse genomes. 
Promoter sequences were extracted from regions 249 nucleotides upstream and 50 nucleotides downstream of the transcription start sites. 

\paragraph{Enhancer} For the enhancer prediction task, we used the dataset from \citep{geng2022} containing DNA sequences classified into strong enhancers, weak enhancers, and non-enhancers. The tasks involved binary classification to distinguish enhancer sequences from non-enhancer sequences and identify specific enhancer types.

\paragraph{Epigenetic Marks} In the epigenetic marks prediction task, we used the dataset from \citep{pham2005,pokholok2005} to predict nucleosome occupancy and modification states in the yeast genome. 
In 10 binary classification tasks, the model had to discriminate between DNA regions that were occupied by histones or not. 
The 10 tasks varied based on the types of histones investigated, including unmodified histones H3 and H4, as well as histones modified by either acetylation (H3K9ac, H3K14ac) or methylation (H3K4me1, H3K4me2, H3K4me3, H3K36me3, H3K79me3).

\paragraph{Splice Site} For the splice site prediction task, DNA sequences from over 100 organisms were used to predict whether the sequences contain donor or acceptor splice sites \citep{scalzitti2021spliceator}.
Donor splice sites denote the beginning of an intron and acceptor splice sites the end of an intron. 
During RNA splicing, these sites are recognized by the spliceosome, a complex molecular machine that enables the removal of introns from the gene.

\paragraph{Preprocessing}
The Nucleotide Transformer study did not provide their exact train-test splits, except for the enhancer dataset.
Therefore, we generated our own train-test splits using a 90:10 ratio. For the promoter dataset, negative samples were not available, and had to be generated following the procedure described by \citep{oubounyt2019deepromoter}. 

\begin{table}
    \small  
    \centering
    \caption{Hyperparameter ranges used to fine-tune {$\sf HyenaDNA$} for all Nucleotide transformer datasets. Exact hyperparameters per dataset can be found in our code \href{https://github.com/HazyResearch/hyena-dna}{repository}.}
    \vspace{10pt}
    \begin{tabular}{lc}
        \toprule
        & {$\sf HyenaDNA$} \\
        \midrule
        Layers & 2 \\
        Width & 256 \\
        Parameters & 1.6M \\
        Optimizer & AdamW \\
        Optimizer momentum & $\beta_1$, $\beta_2$ = 0.9, 0.999 \\
        Training epoch & 100 \\
        Batch size & 256-1024 \\
        Learning rate & 2e-4 to 1e-3 \\
        LR scheduler & Cosine decay \\
        Weight decay (model) & 0-0.2 \\
        Weight decay ({$\sf Hyena$} layers) & 0 \\
        Embed dropout & 0-0.2 \\ 
        Resid dropout & 0-0.2 \\
        Reverse complement aug. & true/false \\
        Sequence lengths & 200-600 \\
        \midrule
    \end{tabular}
    \label{tab:nucleotideTRX_hyperparameter}
\end{table}

\paragraph{Model \& Training} For the architecture, we use a {$\sf HyenaDNA$} model with 2 layers and width 256, and trained on sequences of length 1024. We average across the tokens to obtain a single classification token.
For each task, we replaced the model head and fine-tuned the weights of the entire model (1.6M parameters). In contrast, the Nucleotide Transformer uses a parameter-efficient fine-tuning technique that introduces new weights and fine-tunes only the newly added weights, while keeping the initial model weights frozen, presumably due to its large size of 500M to 2.5B parameters.
The corresponding {$\sf HyenaDNA$} hyperparameter ranges used for training each task are reported in Table~\ref{tab:nucleotideTRX_hyperparameter}.

\subsubsection{Ablations on the Nucleotide Transformer benchmarks}

We perform additional ablations on the Nucleotide Transformer benchmarks to assess the impact of pretraining, as well as attention vs. {$\sf HyenaDNA$}, as shown in shown in Table \ref{tab:nuc-trans-ablations}. We observed that pretraining has a greater effect on the more challenging tasks (and as sequences become longer, shown in \ref{tab:appendix-species-ablation}). On the more challenging tasks (histone marks, datasets starting with “H”), pretraining boosts {$\sf HyenaDNA$} metrics by up to 21 MCC points on H3K4me3. For simpler tasks (with higher baseline scores) such as the splice sites and promoter tasks, the gain was lower (0 to 1 accuracy points), as these were already near saturation in performance. 

\begin{table}[h]
      \small
      \caption{{\bf Pretraining \& Attention ablations on the Nucleotide Transformer (NT) benchmarks.} The Matthews correlation coefficient (MCC) is used as the performance metric for the enhancer and epigenetic marks dataset, and the F1-score is used for the promoter and splice site dataset.
      }
      \label{tab:nuc-trans-ablations}
    \vspace{4mm}
    \centering
\begin{tabular}{lcccc}
            \toprule
            \textsc{Model} 
            & \textsc{NT} & \textsc{GPT} & {$\sf HyenaDNA$} & {$\sf HyenaDNA$} \\
            \textsc{Params}
            & 2.5B & 1.6M & 1.6M & 1.6M \\
            \textsc{Pretrain}
            & yes & yes & yes & no \\
            \midrule
            Enhancer
            & 58.0 & 59.3 & \textbf{62.6} & 58.6 \\
            Enhancer types
            & 47.4 & 51.9 & \textbf{55.7} & 48.4 \\
            H3
            & 81.4 & 75.8 & \textbf{81.7} & 79.9 \\
            H3K4me1
            & 55.9 & 38.7 & \textbf{57.1} & 43.4 \\
            H3K4me2
            & 32.6 & 28.8 & \textbf{53.9}& 34.5 \\
            H3K4me3
            & 42.1 & 28.3 & \textbf{61.2} & 40.2 \\
            H3K9ac
            & 57.5 & 49.2 & \textbf{65.1} & 52.6 \\
            H3K14ac
            & 55.0 & 41.6 & \textbf{66.3} & 48.0 \\
            H3K36me3
            & 63.2 & 47.8 & \textbf{65.3} & 53.4 \\
            H3K79me3
            & 64.2 & 58.9 & \textbf{71.6} & 59.7 \\
            H4
            & \textbf{82.2}	& 77.7 & 79.6 & 79.1 \\
            H4ac
            & 50.1 & 36.4 & \textbf{63.7} & 43.5 \\
            Promoter all
            & \textbf{97.4} & 96.3 & 96.5 & 96.1 \\
            Promoter non-TATA
            & \textbf{97.7} & 96.6 & 96.6 & 96.5\\
            Promoter TATA
            & 96.4 & 96.6 & \textbf{96.7} & 96.1 \\
            Splice acceptor
            & \textbf{99.0} & 97.6 & 96.6 & 96.6 \\
            Splice donor
            & \textbf{98.4} & 98.1 & 97.3 & 96.5 \\
            Splice all
            & \textbf{98.3}	& 98.0 & 97.9 & 97.3 \\
            \bottomrule
        \end{tabular}
\end{table}

%% file: hyenadna/appendix/experiment_subsections/icl.tex
\subsection{In-Context Learning Details}
\label{appendix:icl-details}

\paragraph{Background} A key premise of foundation models is that they are able to learn new tasks with little to no new training data \citep{bommasani2021opportunities}.
Recent advances in language modeling have demonstrated that language foundation models can often adopt the behaviors necessary to perform new tasks \textit{in-context} \citep{brown2020language}.
Here, information about the task that is to be performed, such as examples of respective inputs and targets, are added to the input of the model.
By conditioning their prediction on the provided context, language foundation models are generally able to perform the task without any changes to their parameters.

A key challenge for in-context learning with {$\sf HyenaDNA$} is its limited vocabulary, which is composed of only a few nucleotides, and does not provide any vocabulary for novel downstream tasks, such as class labels.
To explore the potential for in-context learning in genomics, we use two variants of in-context learning, both using a brief tuning phase to introduce {$\sf HyenaDNA$} to the concept of classification with its existing vocabulary.
As a test bed for this exploration, we use $5$ datasets from the GenomicBenchmarks and a {$\sf HyenaDNA$} pretrained on sequences of $160$k length sequences.

In the first experiment, we apply a soft prompting approach \citep{lester2021power} by adding a sequence of tuneable tokens to the input inself.
In the second experiment, we explore a few-shot learning approach \citep{brown2020language} to in-context learning by adding $k$ demonstrations (DNA sequence and its label) for each class of a dataset as input to the model.
To indicate classes, we make use of {$\sf HyenaDNA$}'s existing vocabulary by indicating classes with specific nucleotides.
For binary classification, we indicate classes with the nucleotides "A" and "N", while additionally utilising nucleotide "G" for three-way classification.
During model tuning, we thereby optimise the same next-nucleotide prediction loss as used during pretraining.
See Table \ref{tab:icl-optimisation} for an overview of the optimisation settings.

\begin{figure}[h]
    \centering
    \includegraphics[width=1\linewidth]{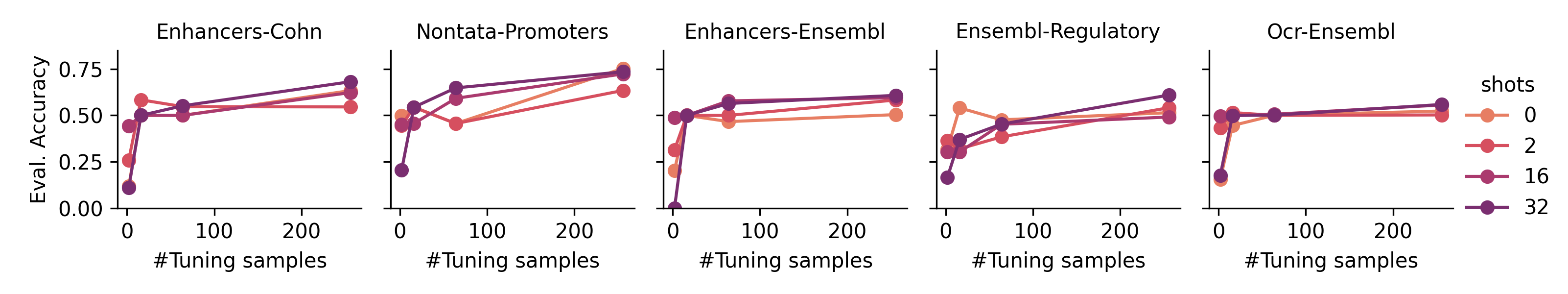}
    \vspace{-2mm}
    \caption{\textbf{Few-shot prompting}: {$\sf HyenaDNA$}'s performance on new tasks generally improves with the number of tuning samples, but is less clear when isolating the number of $k$-shot demonstrations. With less tuning samples, the number of $k$-shot demonstrations do not improve performance. As tuning samples increase, the number of $k$-shot demonstrations start to improve performance.}
\label{fig:fewshot-prompting}
\end{figure}

\paragraph{Soft prompting details} 
For each dataset, we prepend a sequence of $n$ ($2$ to $32$k) learnable tokens $T_e \in \mathbb{R}^{n \times d}$, each of dimension $d$, to the input sequences $X$ of the model: $\{T_e, X, SEP\}$, where "SEP" indicates the separation token. We optimise these tuneable tokens for a maximum of $20$ training epochs on the dataset's training data while keeping all other model parameters fixed. We stop training early if the model's validation loss does not improve for two epochs.
After this tuning phase, we evaluate the model's performance on the dataset's full validation data. 
For an overview of the results of this experiment, see Fig.\ \ref{fig:soft-prompting} of the main text.

\paragraph{Few-shot prompting details}
For each dataset, we prepend a set of $k$ ($0$ to $32$, $0$ indicates regular fine-tuning) examples of each class of a dataset (so-called "shots") to an input sequence:
$$
\texttt{$X$: 
$\{X_1, \text{SEP}, Y_1, \text{SEP}, X_2, \text{SEP}, Y_2, \text{SEP}, X, \text{SEP}\}$}, 
$$

\noindent where $X_i$ indicates an example sequence of class $i$ with label $Y_i$ (exemplified for a two-way classification task).
We tune the model on $n$ ($2$ to $256$) such $k$-shot samples before evaluating its performance on the dataset's full validation data.
For an overview of the results of this experiment, see Fig.\ \ref{fig:fewshot-prompting}.

\begin{table}[h]
    \caption{Optimization settings for in-context learning.}
    \vspace{2mm}
    \label{tab:icl-optimisation}
    \centering
    \begin{tabular}{lcc} \toprule
    {} & \sc{Soft prompting} & \sc{Few-shot prompting} \\ \midrule
    Optimizer & AdamW & AdamW\\
    Optimizer momentum ($\beta_1$, $\beta_2$) & 0.9, 0.999 & 0.9, 0.999 \\
    Learning Rate & 0.001 & 0.0001 \\
    Batch Size & 16 & 2 \\
    Weight Decay (model) & 0 & 0 \\
    Weight Decay ({$\sf Hyena$} layers) & 0 & 0 \\
    Resid dropout & 0 & 0 \\
    Embed dropout & 0.1 & 0.1 \\
    Reverse complement aug. & true & false\\
    LR-schedule & Plateau & - \\
    \bottomrule 
    \end{tabular} 
\end{table}

%% file: hyenadna/appendix/experiment_subsections/chromatin.tex
\subsection{Chromatin Profile Details}
\label{appendix:chromatin-details}

\paragraph{Background} Variations in non-coding regions of the genome account for the majority of disease and other trait-associated single-nucleotide polymorphisms (SNPs). For example, whilst not directly altering the sequence of an encoded protein, a SNP in a non-coding region can affect the expression of downstream genes by inducing a change in the epigenetic state \citep{zaina2010genetics}. Therefore predicting epigenetic markers from a given sequence is an important task in the context of quantifying the functional effects of non-coding variants. Previously DeepSEA \citep{zhou2015predicting}, a deep convolutional sequence model, has been introduced to predict chromatin features directly from non-coding sequences.

\paragraph{Data} The authors of DeepSEA \citep{zhou2015predicting} compiled a dataset of 919 chromatin features from \citep{encode2012integrated} and  \citep{roadmap2015integrative} including 690 TF binding profiles for 160 different TFs, 125 DHS and 104 HM profiles. The original DeepSEA dataset consists of 1000 base pair (bp) sequences from the hg19 human reference genome \citep{church2011modernizing} with corresponding 919-dimension multi-label target vectors. Each label corresponds to the presence/absence of a peak in a given chromatin feature within the central 200 bp region of the sequence. The 400 bp flanking regions of the sequence provide broader contextual information which is beneficial to the task. Training and testing sets are split by chromosome and are strictly non-overlapping. In total, there are 2.2 M training samples and 227,512 samples from chromosomes 8 and 9 are held-out for testing. We use the DeepSEA chromatin profile prediction task to evaluate {$\sf HyenaDNA$} models with varying context window. We use LiftOver \citep{kent2002human} to convert the original DeepSEA dataset to hg38 coordinates and expand flanking regions about the central 200 bp bin symmetrically up to 8000 bp. Approximately 0.5\% of samples are filtered in cases where LiftOver fails or the resulting translated sequence has a different length.

\paragraph{Model} We fine-tune several models consisting of a pretrained {$\sf HyenaDNA$} encoder, a sequence-level pooling layer and a fully-connected decoder to perform multilabel sequence classification. We compare {$\sf HyenaDNA$} against benchmarks set by DeepSEA, a convolutional sequence model, and BigBird \citep{zaheer2020big}, a sparse attention based language model. The authors of BigBird fine-tune on the DeepSEA dataset with input sequences extended to 8000 bp (asymmetrically about the center-point by -5000 and +3000 bp). Notably BigBird utilizes a byte-pair encoding tokenization scheme whereas {$\sf HyenaDNA$} uses a single-character tokenizer and DeepSEA uses one-hot encodings. For the shortest range model (1k), we average across all tokens to perform sequence-level pooling. Whereas in the longer context model (8k) we find that extracting the last token in the sequence as the input to the fully-connected decoder performs better. We also find that for the longer context model using an encoder pretrained on sequences larger than those used in fine-tuning was beneficial. The hyperparameters of the models used in these experiments are shown in Table \ref{tab:chromatin_hyperparams}. Note that we reduced the depth and of models with increasing context window due to limitations on compute cost/time.

\paragraph{Results} The performance of the fine-tuned {$\sf HyenaDNA$} models are summarised in Table \ref{tab:chromatin_profile_aucs}.
We find that the smallest sequence length model (1024 bp) outperforms both DeepSEA and BigBird on TF and DHS prediction. We find that the model pretrained on 32k sequences with only 4 layers and fine-tuned on 8k sequences outperforms BigBird on the long range HM task but suffers from degraded performance on the short range tasks. However, we postulate that this performance loss may be recovered by increasing the depth of the model. We also remark that our models contain 5-30$\times$ fewer parameters compared to DeepSEA and BigBird.

\begin{table}[h]
    \small
    \caption{{\bf Chromatin profile model settings.} {$\sf HyenaDNA$} hyperparameter settings used in the chromatin profile prediction experiments (fine-tuning). 
    }
    \vspace{3mm}
    \centering
    \begin{tabular}{lcc}
    \toprule
    {} & \multicolumn{2}{c}{{$\sf HyenaDNA$}} \\
    \midrule
Sequence length    & 1024                       & 8k \\ 
\midrule
Context window     & 1024                       & 32770                      \\
Width        & 256                        & 256                        \\
Layers             & 8                          & 4                          \\
Pooling method     & Average                       & Last token                 \\
Parameters (M)     & 6.6                        & 3.5                        \\ \midrule
Optimizer          & AdamW                      & AdamW                      \\
Optimizer momentum & $\beta_1,\beta2=0.9,0.999$ & $\beta_1,\beta2=0.9,0.999$ \\
Weight decay (model)      & 0.1                        & 0.1                        \\
Weight decay ({$\sf Hyena$} layers)      & 0                        & 0    
            \\
Embed dropout      & 0.1                        & 0.1                        \\
Learning rate      & 6e-4                       & 6e-4                       \\
Batch size         & 64                         & 64                         \\
Epochs             & 50                         & 50                         \\
\bottomrule
\end{tabular}
        \label{tab:chromatin_hyperparams}
\end{table}

%% file: hyenadna/appendix/experiment_subsections/embeddings.tex
\subsection{Biotype Embeddings Analysis Details}
\label{appendix:embeddings-details}

\paragraph{Background} Sequence embeddings are useful in reducing dimensionality and capturing semantic relationships into fixed length vectors. We analyze pretrained embedding quality from {$\sf HyenaDNA$} and show that it learns biologically informed features. We utilize linear probing, freezing the weights on a pretrained model and attaching a linear classification head to predict biotype sequences. We also use t-SNE to visualize clusterings that emerge from the embeddings.

\paragraph{Data} The Ensembl database \citep{cunningham2022ensembl} is a comprehensive resource for gene and transcript annotations such as biotypes. Ensembl biotypes are a classification system, based on a combination of experimental evidence and computational predictions, that summarises the high-level functional properties of genes and transcripts. For example, biotype classes may annotate whether a gene is protein-coding or encodes a long non-coding RNA; if a gene is a disrupted homologue of a known protein coding gene (pseudogene) and by what mechanism it is produced; or the role of a small non-coding RNA such as post-transcriptional modification of other RNAs in the cell nucleus. We use biotype annotations to qualitatively visualize the clustering of gene embeddings into functional groups. We construct a multi-classification task using the top 10 most frequent biotype annotations as multi-class target labels which we predict from the unsupervised embeddings to assess how well biological function is encoded in the embedding space.

\paragraph{Model \& Training} We use a frozen pretrained {$\sf HyenaDNA$} model consisting of 8 layers and width 256 pretrained on sequences of length 160k. To extract sequence-level embeddings, we average along the sequence dimension in the final encoder layer. For comparison we also construct embeddings using DNABERT (5-mer) and Nucleotide Transformer. We construct embeddings for genes in the Ensembl dataset up to a length of 160k. For genes with sequence lengths exceeding the context window of the encoder, we chunk the sequence and average the embeddings over the chunks. We utilize an XGBoost \citep{chen2016xgboost} classifier to perform the supervised multi-classification task on the embeddings. The hyperparameters used are shown in Table \ref{tab:appendix-biotype}.

\begin{table}[h]{}
    \small
    \caption{{\bf Hyperparameters.} Overview of XGBoost hyperparameters used in biotype multi-classifier. 
    }
    \vspace{3mm}
    \centering
    \begin{tabular}{lc}
    \toprule
    Estimators & 1000 \\
    Max depth & 3 \\
    Learning rate & 0.1 \\
    Objective & softmax\\
\bottomrule
\end{tabular}
        \label{tab:appendix-biotype}
\end{table}

\paragraph{Results} As shown in \ref{tab:embedding_downstream}, {$\sf HyenaDNA$} achieves the highest F1 score on the biotype classification task indicating that its embeddings contain features that are informative of biological function. Notably, {$\sf HyenaDNA$} achieves this using the much smaller embedding space dimension of 256, compared to DNABERT and Nucleotide Transformer, which produce embeddings of dimension 1029 and 1280, respectively.

%% file: hyenadna/appendix/experiment_subsections/species.tex
\subsection{Long-range Species Classification Details}
\label{appendix:species-details}

\begin{table}[h]
    \small  
    \centering
    \caption{Hyperparameter ranges for ultra-long range species classification task. Transformer uses FlashAttention \citep{dao2022flashattention}.}  
    \begin{tabular}{lccccccc}
        \toprule
        & \multicolumn{2}{c}{\sc{Transformer}} & & \multicolumn{4}{c}{{$\sf HyenaDNA$}} \\ 
        \cline{2-3} \cline{5-8} \\
        Layers & 2 & 2 && 2 & 2 & 8 & 8 \\
        Sequence length & 1024 & 32768 && 1024 & 32768 & 250000 & 450000 \\
        Width & 128 & 128 && 128 & 128 & 256 & 256\\
        Parameters (M) & 0.5 & 4.5 && 0.4 & 0.4 & 6.6 & 6.6\\
        Num heads & 8 & 8 && - & - & - & - \\
        Learning rate & $6e^{-5}$ & $6e^{-4}$ && $6e^{-5}$ & $3e^{-4}$ & $6e^{-5}$ & $6e^{-4}$\\
        \midrule
        Optimizer & \multicolumn{6}{c}{AdamW} \\
        Optimizer momentum & \multicolumn{6}{c}{$\beta_1$, $\beta_2$ = 0.9, 0.999} \\
        LR scheduler & \multicolumn{6}{c}{Cosine decay} \\
        Weight decay (model) &\multicolumn{6}{c}{0.1} \\
        Weight decay ({$\sf Hyena$} layers) &\multicolumn{6}{c}{0} \\
        Embed dropout & \multicolumn{6}{c}{0.1} \\
        Resid dropout & \multicolumn{6}{c}{0} \\
        Batch size & \multicolumn{6}{c}{128 - 256} \\
        Training epoch & \multicolumn{6}{c}{200} \\
        Reverse complement aug. & \multicolumn{6}{c}{False} \\
        \bottomrule
    \end{tabular}
    \label{tab:appendix-species-classification}
\end{table}

\paragraph{Background} Given a genetic sequence randomly sampled from a set of different species, successful identification of the source species requires a model to learn a distinct mutational profile for each species. The more locations for discriminative mutations a model can consider, the more successful it should be at this task. We can arbitrarily tune this task's difficulty by including a higher number of species or increasing the evolutionary similarity of the included species, and thus it represents a helpful setting for measuring long context reasoning abilities for DNA sequence models.

\paragraph{Data} We select five species for this task: human (\textit{homo sapien}), lemur (\textit{lemur catta}), mouse (\textit{mus musculus}), pig (\textit{sus scrofa}), and hippo (\textit{hippopotamus amphibius}). We hold out four chromosomes from each species (chromosome numbers 1, 3, 12, and 13) for evaluation, and use the rest of each species' chromosomes for training.

\paragraph{Model} We compare {$\sf HyenaDNA$} against a baseline Transformer, which uses Flash Attention \citep{dao2022flashattention} in the mixing layer instead of a Hyena operator. We use 2 and 8 layer models, depending on sequence length. For {$\sf HyenaDNA$}, we train on sequence lengths of 1k, 32k, 250k, 450k and 1M. For Transformer, we limit sequence lengths to 1k and 32k due to the quadratic increase in training time, making training infeasible on our hardware. See Table \ref{tab:appendix-species-classification} for model sizes and hyperparamters.

\paragraph{Training} 

We use pretrained models from \ref{experiments:pretraining}, trained on various lengths between 1k to 1M nucleotides, and fine-tune them using a linear decoder head. We either pool across all tokens (1k and 32k models) or use the last token for classification (250k - 1M models). We randomly sample a \textit{(species, chromosome, sequence start, sequence end)} tuple at each training step, with uniform probability across all species and non-held-out chromosomes. If a sequence's starting location on a chromosome is such that the end of that sequence would exceed the length of the chromosome, then we pad the sequence with N's to its full intended length. For evaluation, we randomly sample a \textit{(species, chromosome, sequence start, sequence end)} tuple from our held-out evaluation set of chromosomes, and record the overall Top-1 5-way accuracy of our model (i.e. fraction of sequences correctly classified).

At sequence length 450k, we use the sequence length warm-up scheduler described in \ref{warmup} on {$\sf HyenaDNA$}. This involves gradually increasing the length of sequences fed to the model during fine-tuning from 1k to 450k. We observe better convergence and higher overall peak accuracy with this strategy, as shown in \ref{fig:warmup_acc}.

\paragraph{Pretraining ablation}

\begin{table}[h]{}
    \small
    \caption{{\bf Pretraining vs scratch on 5-way species classification.} Top 1\% accuracy for $\sf{HyenaDNA}$ by sequence length.
    }
    \vspace{3mm}
    \centering
    \begin{tabular}{lcc}
    \toprule
    {} & \multicolumn{2}{c}{{$\sf HyenaDNA$}} \\
    \midrule
    \sc{Length} & \sc{scratch} & \sc{Pretrained} \\
    1k & 53.9 & 61.1 \\
    32k & 70.7 & 93.4 \\
    250k & 65.7 & 97.9 \\
    450k & 71.4 & 99.4 \\
\bottomrule
\end{tabular}
        \label{tab:appendix-species-ablation}
\end{table}

For species classification, pretraining becomes more important for longer sequences. This is in-line with our observation that for harder tasks (including longer sequences), pretraining becomes more important. At sequence length 250k and 450k, the scratch vs. pretraining gap is 30+ accuracy points.